\documentclass[10pt,twocolumn,letterpaper]{article}

\usepackage[pagenumbers]{cvpr}

\usepackage{amsmath}
\usepackage{amssymb}
\usepackage{graphicx}
\usepackage{booktabs}
\usepackage{multirow}
\usepackage{array}
\usepackage{xcolor}
\usepackage{colortbl}
\usepackage{algorithm}
\usepackage{algpseudocode}

\definecolor{cvprblue}{rgb}{0.21,0.49,0.74}
\usepackage[pagebackref,breaklinks,colorlinks,allcolors=cvprblue]{hyperref}

\def\paperID{arXiv}
\def\confName{CVPR}
\def\confYear{2026}

\title{DynTrace: Tracking Dynamic Object Evidence for 4D Spatio-Temporal Reasoning in MLLMs}

\author{
Rongxin Gao$^{1,\dagger}$ \quad
Yuzhi Huang$^{2,\dagger,\ddagger}$ \quad
Dongxuan Liu$^{1,\dagger}$ \quad
Chu Li$^{3}$ \quad
Zhenye Wang$^{3}$ \quad
Jie Wu$^{2}$ \\
Shuzhao Xie$^{2}$ \quad
Jingyan Jiang$^{4,*}$ \quad
Xinghao Ding$^{1}$ \quad
Xiaotong Tu$^{1,*}$ \quad
Yue Huang$^{1}$ \\
$^{1}$Xiamen University \quad
$^{2}$Shenzhen International Graduate School, Tsinghua University  \\
$^{3}$China University of Mining and Technology \quad
$^{4}$Shenzhen Technology University \\
$^{\dagger}$Equal contribution.\quad
$^{*}$Corresponding authors.\quad
$^{\ddagger}$Project lead.
}

\definecolor{sectiongray}{RGB}{240,240,240}
\definecolor{headergray}{RGB}{230,230,230}
\definecolor{darkgreen}{RGB}{0,150,0}
\definecolor{darkred}{RGB}{200,0,0}
\definecolor{lightblue}{RGB}{200,220,255}
\definecolor{orange}{RGB}{255,200,100}
\definecolor{lightyellow}{RGB}{255,255,200}

\begin{document}
\maketitle
\begingroup
\renewcommand\thefootnote{}
\footnotetext{
\textbf{Project page} :\href{https://dyntrace.github.io/}{\texttt{https://dyntrace.github.io/}}
}
\endgroup
\begin{abstract}
\label{sec:abs}
4D spatio-temporal reasoning, jointly modeling 3D spatial structure and temporal evolution, is essential for understanding dynamic worlds and enabling embodied interaction. While current Multimodal Large Language Models (MLLMs) show strong capabilities in static scene understanding and coarse-grained 4D tasks, they still have notable limitations in continuous dynamic scene perception, especially in tracking dynamic object evidence for coherent 4D spatio-temporal reasoning. This shortcoming stems mainly from relying on sparse frame-level observations, fragmenting continuous dynamic cues and leaving models unable to disentangle genuine object dynamics from camera-induced apparent motion. Inspired by humans tracking dynamic cues while compensating for viewpoint changes, we propose DynTrace, a training-free framework for 4D spatio-temporal reasoning with two complementary components. Dynamic Trajectory Visualization (DTV) reprojects world-coordinate trajectories onto the image plane, providing geometry-informed visual priors that disentangle genuine object dynamics from camera-induced apparent motion. Meanwhile, the Dynamic Trace Token (DT-Token), organized into a Dynamic Trace Graph (DTG), tracks object-level dynamic cues, trace evolution, and key moments, maintaining continuous dynamic object evidence for coherent 4D reasoning. Together, these two components equip MLLMs with continuously tracked dynamic object evidence, grounded in geometry-informed visual priors and structured spatio-temporal traces. DynTrace consistently improves open-source MLLMs, achieving state-of-the-art results on Dyn-Bench, VLM4D, and DSI-Bench, validating the importance of tracking dynamic object evidence for robust 4D spatio-temporal reasoning.
\begin{figure*}[t]
  \centering
  \includegraphics[width=\textwidth]{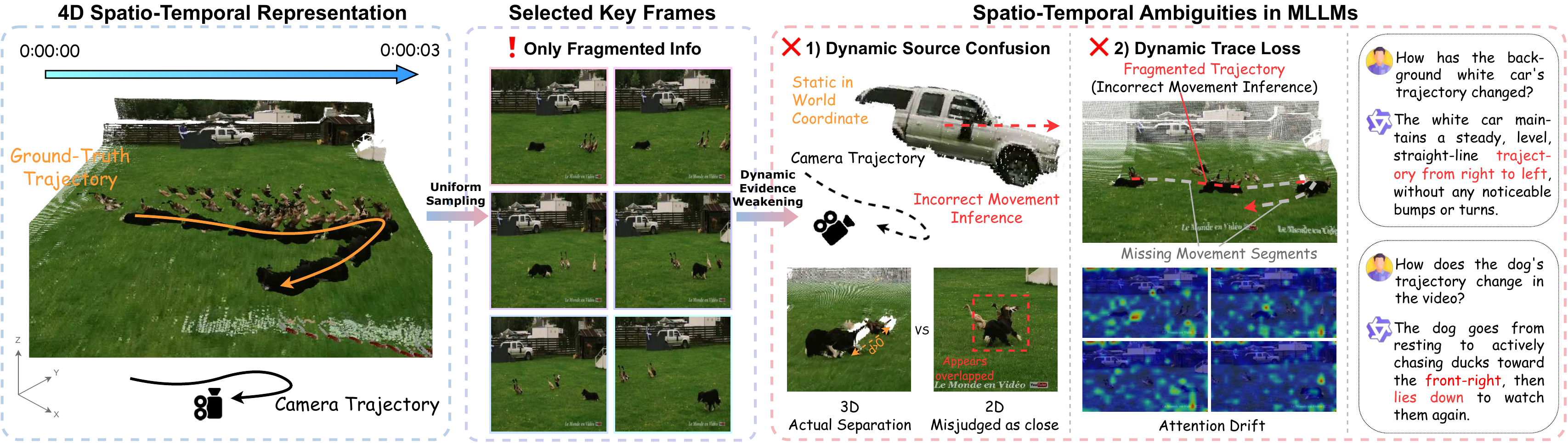}
  \caption{Failure modes of MLLMs in 4D spatio-temporal reasoning. Uniform sampling yields fragmented key frames that weaken dynamic evidence, causing two major errors: (1) Dynamic Source Confusion, where camera trajectories cause incorrect movement inference, misjudging true 3D separation as close in 2D; (2) Dynamic Trace Loss, where ground-truth trajectories degrade into fragmented trajectories with missing segments and attention drift.}
  \label{fig:intro}
\end{figure*}
\end{abstract}
\vspace{-2em}
\section{Introduction}
\label{sec:intro}

4D spatio-temporal reasoning jointly models 3D space and time, requiring models to capture scene structure and reason about how objects move, interact, and evolve. This capability is essential for embodied intelligence, robotic interaction, and open-world understanding, which require sustained perception of scene geometry and dynamic object behaviors~\cite{navid-4d,st-vla,sohn2025r4,sohn2025snow,zhou2025llava4d,yang20264drgpt,zhou2025learning, Mehta2025LargeLM, Chen2025Exploring, Dong2025TowardEI,huang2026robostream,wen2025dynamicverse}.

Recent Multimodal Large Language Models (MLLMs) excel in static scene perception and understanding~\cite{zhu2025internvl3,Qwen3-Omni,LLaVA-OneVision-1.5,
damonlpsg2025videollama3, wang2025internvideo, Qwen3-VL,
wang2025internvl3_5, xu2025learning,
meng2025openo3, yuan2025videorefer, yuan2025pixelrefer}, yet remain limited in 4D spatio-temporal reasoning owing to their emphasis on object recognition and coarse-grained video question answering, which leaves them falling short in fine-grained spatial relation inference and temporal dynamics modeling. In other words, strong appearance understanding does not automatically translate into reliable reasoning over continuously changing scenes and evolving interactions. Motivated by this gap, recent 4D benchmarks push evaluation beyond these tasks toward unified reasoning over spatial relations, object size, scene geometry, and temporal dynamics~\cite{wang2025spatial4d,zhu20254dbenchbenchmarkingmultimodallarge,li2025sti,wu2026st,lin2025mmsi,yang2024think}. 
However, continuous dynamics perception, which requires distinguishing true object movement from camera ego-motion while tracking multiple entities' trajectories and relation evolution, remains an underexplored dimension, particularly in how models preserve object-centered evidence as motion unfolds. The bottleneck thus lies not only in a limited video context, but in failing to continuously track dynamic object evidence across changing viewpoints and long horizons. 

Human 4D cognition offers a design principle to close this gap: rather than treating ego-motion compensation and entity tracking as separate problems, humans inherently couple these tasks, disentangling camera motion from object motion while maintaining sustained attention to salient interacting entities across time~\cite{zhou2025feature4x, luiten2023dynamic, lei2024mosca}. Dyn-Bench~\cite{huang2026thinking} empirically validates this coupled mechanism, showing that effective dynamic-scene reasoning depends on jointly understanding camera movement and tracking evolving objects in a unified and object-centered manner. Motivated by these findings, recent efforts have enhanced MLLMs' spatio-temporal reasoning by adding 3D or spatial supervision in training~\cite{spacer, batra2025spatialthinker, spatial-ssrl,yang20264drgpt}, injecting geometric cues such as metric depth and pose~\cite{depthanything3, huang2025vipe, chen2025gca,wu2025spatialmllm}, and augmenting inference with memory for frame-level observations or intermediate reasoning states~\cite{wang2024videotree, liu2025bolt, tang2025adaptive, huang2026robostream,song2024moviechat,he2024ma}.

Despite these efforts, current methods
~\cite{spacer,spatial-ssrl,batra2025spatialthinker,chen2025gca,zheng2025learning,yang20264drgpt,yin2026mllm4d,wu2025spatialmllm}
still struggle to track dynamic object evidence reliably over time, as shown in Fig.~\ref{fig:intro}, owing to two prevalent failure modes. (i)~\textit{Dynamic Source Confusion}: ego-motion-aware perception requires disentangling camera motion from genuine object movement, yet even methods injecting geometric cues such as depth or pose do not model ego-motion as a separate factor explicitly. When the camera moves amid multi-entity interactions, models often attribute ego-motion-induced apparent motion to objects, misidentify which entities truly move, and distort relational judgments, making tracked dynamics unreliable. (ii)~\textit{Dynamic Trace Loss}: continuous object tracking requires sufficiently dense temporal coverage to preserve full motion trajectories, yet current strategies rely on sparse frame sampling, temporal token compression, or clip-level summaries rather than continuous object-centric traces. Consequently, dynamic objects' temporal evidence chain fragments, and model attention drifts toward unstable background changes over truly salient motion.
Moreover, these modes mutually reinforce each other, as fragmented traces complicate ego-motion factoring and source-confused evidence degrades tracking consistency, which ultimately prevents MLLMs from maintaining the continuous dynamic object evidence required for downstream 4D reasoning in complex scenes.

To bridge this gap, we propose \textbf{DynTrace}, a training-free framework coupling geometry-informed visual priors with structured spatio-temporal traces to provide MLLMs with explicit dynamic object evidence for robust 4D reasoning. DynTrace constructs temporally consistent dynamic instances and employs two synergistic components. First, \textit{Dynamic Trajectory Visualization (DTV)} reprojects motion trajectories onto the image plane to generate visual priors, separating true object movement from camera-induced motion to mitigate \textit{Dynamic Source Confusion}. Second, \textit{Dynamic Trace Graph (DTG)} organizes the \textit{Dynamic Trace Token (DT-Token)}, encoding dynamic cues, trace evolution, and key moments over time, into a queryable graph that explicitly preserves long-range traces and relation changes to directly resolve \textit{Dynamic Trace Loss}. By integrating DTV and DTG, DynTrace explicitly realizes this coupled mechanism during inference, ensuring that disentangling camera motion and tracking interacting entities reinforce each other. Evaluations on Dyn-Bench~\cite{huang2026thinking}, VLM4D~\cite{zhou2025vlm4d}, and DSI-Bench~\cite{zhang2025dsibench} demonstrate that DynTrace consistently improves open-source MLLMs across diverse and challenging 4D spatio-temporal reasoning tasks. In summary, our contributions are as follows:
\begin{itemize}
\item We introduce \textbf{DynTrace}, a training-free framework for 4D spatio-temporal reasoning in MLLMs. By explicitly constructing continuously tracked dynamic object evidence, it improves understanding of complex dynamic scenes.

\item We design two complementary dynamic priors from visual and textual perspectives: \textit{DTV}, which disentangles genuine object dynamics from camera-induced apparent motion through trajectory reprojection, and \textit{DTG}, which organizes the \textit{DT-Token} into a queryable graph preserving long-horizon dynamic traces for coherent reasoning.

\item Evaluations on Dyn-Bench, VLM4D, and DSI-Bench demonstrate that DynTrace consistently improves open-source MLLMs across diverse 4D reasoning tasks, validating that tracking dynamic object evidence is important for robust 4D spatio-temporal reasoning.
\end{itemize}
\section{Related Work}
\subsection{Video Understanding MLLMs}
Video understanding MLLMs aim to retain query-relevant events, object states, and interactions over long and continuously changing videos. Existing methods mainly enhance this capability from two directions. The first direction focuses on representation efficiency, including key-frame selection, clip scheduling, temporal token compression, and hierarchical aggregation, so that videos can be processed within a limited context budget~\cite{tang2025adaptive, wang2025internvideo, jiang2025storm, li2025videochat, yuan2026unicomp, wu2026a, shen2024longvu}. The second focuses on evidence organization, such as turning videos into document-style intermediates, building hierarchical summaries over events and objects, or retrieving relevant clips on demand during inference~\cite{wang2024videotree, yuan2025think, liu2025bolt, wang2025videoitg}. Query-guided filtering is also widely used to remove irrelevant frames or regions before the final MLLM reasoning stage~\cite{tang2025adaptive, li2025videochat, yuan2025think}. While effective for retaining high-level semantics under context constraints, these methods still mainly \emph{select}, \emph{compress}, or \emph{summarize} video content before reasoning, making them less suited to preserving continuously tracked object-centered dynamics. Consequently, semantically selected evidence struggles to disentangle genuine object dynamics from camera-induced motion, leaving \textit{Dynamic Source Confusion} unresolved. Furthermore, reducing motion to sparse temporal tokens or retrieved snippets fragments the evidence chain, causing \textit{Dynamic Trace Loss}.

\begin{figure*}[t]
  \centering
  \includegraphics[width=\textwidth]{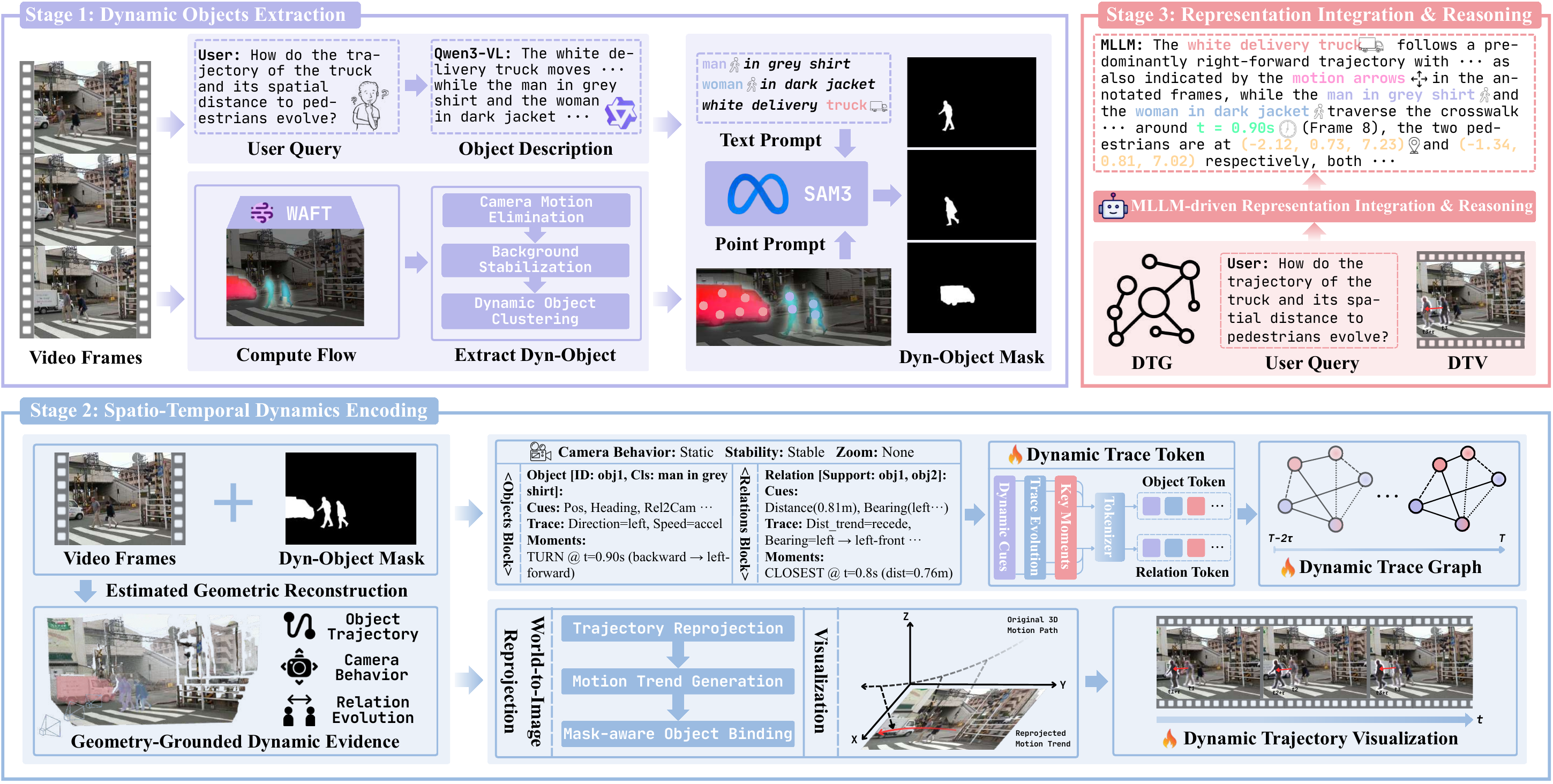}
  \caption{Overview of DynTrace. Given a video and a language query, DynTrace follows three stages: (1) Dynamic Objects Extraction identifies query-relevant and independently moving instances, producing temporally consistent dynamic masks; (2) Spatio-Temporal Dynamics Encoding lifts tracked instances into a shared world frame to reconstruct Geometry-Grounded Dynamic Evidence, including Object Trajectory, Camera Behavior, and Relation Evolution, from which it derives DTV through World-to-Image Reprojection and converts Dynamic Cues, Trace Evolution, and Key Moments into object-side and relation-side DT-Tokens before organizing them into a DTG; and (3) Representation Integration \& Reasoning feeds DTV, DTG, and the query into the target MLLM for 4D spatio-temporal reasoning.}
  \label{fig:pipeline}
\end{figure*}

\subsection{4D Spatio-temporal Reasoning}
Recent work on 4D spatio-temporal reasoning further introduces geometric structure, world representations, and external tools into MLLMs. Depth, optical flow, camera pose, and 3D reconstruction are used to strengthen spatial grounding and temporal consistency~\cite{zhou2025vlm4d, zhou2025learning, zhou2025llava4d, yang20264drgpt, huang2025vipe, lu2024llavamr, pan2025diff4splat, li20244k4dgen, lin2025movies}. Structured memories and world models organize object states, relations, and temporal indices into queryable records for longer-horizon reasoning~\cite{huang2026thinking, sohn2025r4, sohn2025snow, wen2025dynamicverse}. Tool-augmented pipelines also offload geometry estimation, trajectory recovery, and constraint solving to external modules before feeding results back to the language model~\cite{fan2025tool, chen2025gca}. These studies move beyond plain frame aggregation and push MLLMs toward richer 4D reasoning, but they mainly use geometry, world states, and tools to enhance grounding, state organization, or intermediate computation rather than to expose continuously tracked dynamic evidence directly to the MLLM. 
However, failing to explicitly decouple object from camera motion perpetuates \textit{Dynamic Source Confusion}, particularly within ego-centric and multi-entity scenarios. Furthermore, abstracting evidence into sparse states or tool outputs fractures the continuous motion chain, thereby inducing \textit{Dynamic Trace Loss}. \textbf{DynTrace} mitigates these critical limitations. Rather than compelling the backbone toward implicit dynamic inference, it integrates DTV and DTG at inference time. This endows the architecture with geometry-informed visual priors and a compact dynamic trace that maintains temporal explicitness.
\section{Method}
\label{sec:method}
Current MLLMs still struggle to preserve dynamic object evidence for 4D reasoning tasks. In practice, they must infer both motion source and spatio-temporal traces from sparse frames, which directly induces \textit{Dynamic Source Confusion} and \textit{Dynamic Trace Loss}. DynTrace addresses this gap by converting raw video into explicit dynamic evidence before final MLLM reasoning.

As shown in Fig.~\ref{fig:pipeline}, given a video $\mathcal{V}=\{I_t\}_{t=1}^{T}$ and a language query $q$, DynTrace comprises three stages. \textit{Dynamic Objects Extraction} localizes query-relevant moving instances and produces temporally consistent dynamic masks. \textit{Spatio-Temporal Dynamics Encoding} reconstructs \textit{Geometry-Grounded Dynamic Evidence} from the original video and the masks, then derives the Dynamic Trajectory Visualization (DTV) $\mathcal{D}$ and the Dynamic Trace Graph (DTG) $\mathcal{G}$. \textit{Representation Integration \& Reasoning} finally feeds $\mathcal{D}$, $\mathcal{G}$, and $q$ into the target MLLM. DTV provides geometry-informed visual priors on the image plane, while DTG organizes DT-Tokens that preserve dynamic cues, trace evolution, and key moments. Consequently, the framework shifts 4D reasoning from sparse whole-frame impressions to persistent dynamic object evidence.

\subsection{Dynamic Objects Extraction}
The first stage extracts dynamic objects that are both query-relevant and independently moving from the video. We first use Qwen3-VL-8B to decompose $q$ into a set of visual entity descriptions:
\begin{equation}
\mathcal{E}={\mathrm{MLLM}}(\mathcal{V},q)=\{e_1,e_2,\cdots,e_N\},
\end{equation}
where $N$ is the number of extracted entities and each $e_i$ serves as a text prompt for subsequent segmentation and tracking of candidate objects. This semantic filtering matters because dynamic videos often contain many moving regions, but only a small subset is actually query-relevant. By pruning the candidate set early, DynTrace keeps later stages explicitly focused on objects carrying the required dynamic evidence.

We next estimate dense optical flow with WAFT (Warping-Alone Field Transforms)~\cite{wang2025waft} on consecutive frames and decouple scene-wide drift from independently moving objects. This step plays the role of \textit{Camera Motion Elimination}, because it explicitly estimates the dominant background flow caused by camera motion and subtracts it from the full motion field:
\begin{equation}
\mathbf{F}_t=f_{\mathrm{Flow}}(I_t,I_{t+1}), \quad
\hat{\mathbf{F}}^{\mathrm{bg}}_t=f_{\mathrm{BG}}(\mathbf{F}_t),
\end{equation}
where $f_{\mathrm{Flow}}$ denotes the WAFT flow estimator, $\mathbf{F}_t$ is the dense flow, $f_{\mathrm{BG}}$ extracts the dominant background component, and $\hat{\mathbf{F}}^{\mathrm{bg}}_t$ is the estimated background flow. The residual motion then follows:
\begin{equation}
\mathbf{F}^{\mathrm{res}}_t=\mathbf{F}_t-\hat{\mathbf{F}}^{\mathrm{bg}}_t,
\end{equation}
which suppresses scene-wide drift and stabilizes the remaining motion field. In this sense, \textit{Background Stabilization} is achieved by retaining only the motion that cannot be explained by the camera ego-motion, providing a cleaner basis for instance extraction in videos with complex viewpoint changes.

We then perform \textit{Dynamic Object Clustering} by grouping residual motion points into dynamic-object proposals and point seeds $\mathcal{S}$. This aggregates scattered motion cues into object-level hypotheses, enabling the subsequent tracking stage to operate on coherent entities rather than isolated pixels. Finally, we feed text prompts $\mathcal{E}$ and point prompts $\mathcal{S}$ into SAM3 (Segment Anything Model 3)~\cite{meta2025sam3} to obtain temporally consistent dynamic masks:
\begin{equation}
\mathcal{M}=f_{\mathrm{Seg}}(\mathcal{V},\mathcal{E},\mathcal{S})=\{M_i^t\mid i=1,\dots,N_t,\; t=1,\dots,T\},
\end{equation}
where $f_{\mathrm{Seg}}$ denotes the SAM3-based segmentation-and-tracking function, $M_i^t$ is the dynamic mask of object $i$ at frame $t$, and $N_t$ is the number of tracked dynamic objects at time $t$. At the end of this stage, DynTrace obtains temporally consistent dynamic masks aligned with both the query and the underlying dynamic instances, establishing the object-centric basis for subsequent geometric reconstruction and structured tracing.

\begin{table*}[t]
\centering
\caption{Comparison on Dyn-Bench across nine dynamic reasoning categories. The enhanced rows apply DynTrace at inference time. Bold and underlined values denote the best and second-best results, respectively.}
\label{tab:dynbench_results}
\setlength{\tabcolsep}{2.5pt}
\renewcommand{\arraystretch}{1.0}
\resizebox{\textwidth}{!}{%
\begin{tabular}{l|c|ccccccccc}
\toprule
\multirow{1.5}{*}{\textbf{Method}} &
\multirow{1.5}{*}{\textbf{Avg.}} &
\textbf{\scriptsize \shortstack{Act. \& \\  Obj. Desc.}} &
\textbf{\scriptsize \shortstack{Move. \& \\  Temp. Dyn.}} &
\textbf{\scriptsize \shortstack{Spatial Rel. \\ \& Change}} &
\textbf{\scriptsize \shortstack{Mov. Patterns \\ \& Traj.}} &
\textbf{\scriptsize \shortstack{Spatial Rel. \\ \& Comp.}} &
\textbf{\scriptsize \shortstack{Scene Focus \\ \& Dyn.}} &
\textbf{\scriptsize \shortstack{Cam. Motion \\ \& Orient.}} &
\textbf{\scriptsize \shortstack{Cam-Obj. \\ Interaction}} &
\textbf{\scriptsize \shortstack{Temp. \& \\ Visual Change}} \\
& &
\multicolumn{3}{c}{\cellcolor{orange!25}\textbf{Inter-Object}} &
\multicolumn{3}{c}{\cellcolor{yellow!25}\textbf{Object-Scene}} &
\multicolumn{3}{c}{\cellcolor{blue!15}\textbf{Camera-Object}}\\
\midrule
\rowcolor{gray!10}\multicolumn{11}{l}{\textit{Spatial MLLMs}} \\
SpaceR-7B~\cite{spacer} & 56.5 & 66.6 & 49.2 & 52.7 & 72.2 & 67.8 & 78.2 & 50.3 & 40.0 & 55.5 \\
VST-7B-RL~\cite{vst} & 55.7 & 68.6 & 48.4 & 51.9 & 73.0 & 70.7 & 79.4 & 45.1 & 39.1 & 52.9 \\
Spatial-SSRL-7B~\cite{spatial-ssrl} & 45.9 & 54.5 & 40.0 & 48.1 & 68.5 & 65.9 & 73.8 & 35.8 & 36.7 & 37.7 \\
SpatialReasoner~\cite{ma2025spatialreasoner} & 54.5 & 63.2 & 44.1 & 50.4 & 68.2 & 64.2 & 74.0 & 48.0 & 44.6 & 54.0 \\
SpatialThinker-7B~\cite{batra2025spatialthinker} & 53.7 & 63.2 & 40.7 & 46.5 & 70.6 & 67.3 & 77.5 & 46.0 & 42.8 & 51.5 \\
MLLM-4D-8B~\cite{yin2026mllm4d} & 56.4 & 63.6 & 44.5 & 47.4 & 64.3 & 61.4 & 70.6 & 52.4 & 52.3 & 63.7 \\
LLaVA-ST-7B~\cite{li2025llavast} & 50.2 & 56.3 & 40.3 & 48.7 & 62.3 & 59.8 & 70.1 & 47.8 & 38.9 & 46.8 \\
\midrule
\rowcolor{gray!10}\multicolumn{11}{l}{\textit{Video MLLMs}} \\
LLaVA-OV-1.5-8B~\cite{LLaVA-OneVision-1.5} & 53.8 & 60.9 & 47.7 & 53.4 & 74.4 & 69.6 & 75.4 & 41.0 & 37.0 & 51.6 \\
Videorefer-7B~\cite{yuan2025videorefer} & 56.1 & 65.2 & 50.9 & 56.5 & 73.2 & 72.4 & 79.0 & 36.0 & 43.0 & 56.1 \\
InternVideo2.5-Chat-8B~\cite{wang2025internvideo} & 54.2 & 67.9 & 48.5 & 46.3 & 70.1 & 65.8 & 76.5 & 42.0 & 42.9 & 52.1 \\
VideoLLaMA3-7B~\cite{damonlpsg2025videollama3} & 54.3 & 63.2 & 40.7 & 46.5 & 70.6 & 67.3 & 77.5 & 46.0 & 42.8 & 51.5 \\
\midrule
\rowcolor{gray!10}\multicolumn{11}{l}{\textit{Training-free MLLMs}} \\
See\&Trek~\cite{li2025seetrek} & 51.5 & 67.2 & 44.8 & 49.5 & 69.5 & 64.7 & 72.9 & 47.2 & 34.5 & 41.0 \\
GSM~\cite{zhou2025learning} & 55.3 & 63.9 & 49.3 & 49.9 & 73.3 & 70.8 & 79.2 & 48.5 & 36.7 & 50.8 \\
\midrule
\rowcolor{gray!10}\multicolumn{11}{l}{\textit{Ours}} \\
Qwen3-VL-8B-Instruct~\cite{Qwen3-VL} & 60.8 & 70.8 & 52.6 & 53.6 & 75.0 & 71.2 & 79.0 & 54.3 & 51.4 & 59.7 \\
\rowcolor{blue!5}\quad \textbf{+DynTrace} & \underline{65.8} \small{\textcolor{darkred}{+5.0}} & \underline{79.7} \small{\textcolor{darkred}{+8.9}} & \underline{57.1} \small{\textcolor{darkred}{+4.5}} & \underline{62.2} \small{\textcolor{darkred}{+8.6}} & \textbf{80.7} \small{\textcolor{darkred}{+5.7}} & \underline{75.3} \small{\textcolor{darkred}{+4.1}} & \textbf{85.6} \small{\textcolor{darkred}{+6.6}} & \textbf{58.9} \small{\textcolor{darkred}{+4.6}} & 52.6 \small{\textcolor{darkred}{+1.2}} & \underline{64.4} \small{\textcolor{darkred}{+4.7}} \\
Qwen3-VL-32B-Instruct~\cite{Qwen3-VL} & 62.4 & 71.4 & 54.6 & 56.1 & 75.3 & 74.4 & 79.8 & 55.9 & \underline{53.4} & 58.4 \\
\rowcolor{blue!5}\quad \textbf{+DynTrace} & \textbf{66.9} \small{\textcolor{darkred}{+4.5}} & \textbf{82.4} \small{\textcolor{darkred}{+11.0}} & \textbf{61.8} \small{\textcolor{darkred}{+7.2}} & \textbf{66.2} \small{\textcolor{darkred}{+10.1}} & \textbf{80.7} \small{\textcolor{darkred}{+5.4}} & \textbf{75.7} \small{\textcolor{darkred}{+1.3}} & \underline{84.5} \small{\textcolor{darkred}{+4.7}} & \underline{56.7} \small{\textcolor{darkred}{+0.8}} & \textbf{53.9} \small{\textcolor{darkred}{+0.5}} & \textbf{65.6} \small{\textcolor{darkred}{+7.2}} \\
InternVL3.5-8B~\cite{wang2025internvl3_5} & 53.2 & 69.2 & 44.4 & 47.4 & 66.2 & 63.7 & 71.9 & 44.3 & 44.8 & 49.3 \\
\rowcolor{blue!5}\quad \textbf{+DynTrace} & 58.0 \small{\textcolor{darkred}{+4.8}} & 73.9 \small{\textcolor{darkred}{+4.7}} & 50.7 \small{\textcolor{darkred}{+6.3}} & 52.6 \small{\textcolor{darkred}{+5.2}} & 70.8 \small{\textcolor{darkred}{+4.6}} & 68.0 \small{\textcolor{darkred}{+4.3}} & 80.2 \small{\textcolor{darkred}{+8.3}} & 47.7 \small{\textcolor{darkred}{+3.4}} & 48.3 \small{\textcolor{darkred}{+3.5}} & 53.8 \small{\textcolor{darkred}{+4.5}} \\
InternVL3.5-14B~\cite{wang2025internvl3_5} & 56.0 & 72.3 & 49.6 & 47.1 & 70.6 & 68.2 & 75.6 & 48.8 & 47.0 & 46.2 \\
\rowcolor{blue!5}\quad \textbf{+DynTrace} & 60.3 \small{\textcolor{darkred}{+4.3}} & 74.6 \small{\textcolor{darkred}{+2.3}} & 52.7 \small{\textcolor{darkred}{+3.1}} & 59.8 \small{\textcolor{darkred}{+12.7}} & \underline{75.5} \small{\textcolor{darkred}{+4.9}} & 72.0 \small{\textcolor{darkred}{+3.8}} & 80.5 \small{\textcolor{darkred}{+4.9}} & 51.1 \small{\textcolor{darkred}{+2.3}} & 48.9 \small{\textcolor{darkred}{+1.9}} & 53.1 \small{\textcolor{darkred}{+6.9}} \\
\bottomrule
\end{tabular}%
}
\end{table*}

\subsection{Spatio-Temporal Dynamics Encoding}
Given the original video $\mathcal{V}$ and dynamic masks $\mathcal{M}$, the second stage transforms tracked instances into metric dynamic evidence and then derives DTV and DTG from it. We first employ DA3 (Depth Anything 3)~\cite{depthanything3} on each frame to estimate depth and camera pose. Using the centroid $(u_i^t,v_i^t)$ of mask $M_i^t$, the median mask depth $z_i^t$, the intrinsic matrix $\mathbf{K}_t$, and the camera-to-world transform $\mathbf{T}_{c\rightarrow w}^t$, the world position of object $i$ at time $t$ is recovered by
\begin{equation}
\mathbf{p}_i^t=\mathbf{T}_{c\rightarrow w}^t\left(z_i^t\mathbf{K}_t^{-1}[u_i^t,v_i^t,1]^\top\right),
\end{equation}
where $\mathbf{T}_{c\rightarrow w}^t(\cdot)$ applies the estimated camera-to-world rigid transform to the 3D point in camera coordinates. This step lifts every tracked instance to a shared world frame, ensuring that motion is explicitly disentangled from the camera viewpoint. From $\{\mathbf{p}_i^t\}_{t=1}^{T}$, we obtain the continuous \textit{Object Trajectory} of each dynamic object. From the DA3 pose sequence, we summarize \textit{Camera Behavior}, including whether the camera remains stable, translates, or changes zoom. For each object pair $(i,j)$, we further compute their spatial metric relation over time through distance $d_{ij}^t=\|\mathbf{p}_i^t-\mathbf{p}_j^t\|_2$ and relative bearing $b_{ij}^t=f_{\mathrm{bear}}(\mathbf{p}_j^t-\mathbf{p}_i^t)$, where $f_{\mathrm{bear}}(\cdot)$ maps the relative displacement to a bearing descriptor; together, they define \textit{Relation Evolution}. These three streams provide the common basis for both the visual branch and the structured textual branch.
\paragraph{Dynamic Trajectory Visualization (DTV)}
DTV serves as the visual branch. We define the DTV collection as $\mathcal{D}=\{D_i^t\mid i=1,\dots,N_t,\; t=1,\dots,T\}$, where $D_i^t$ visually tracks object $i$ at frame $t$. For each dynamic mask $M_i^t$, let $(u_i^t,v_i^t)$ be its image-plane centroid and $\Omega_i^t$ be its rendering support. Furthermore, let $\hat{\mathbf{v}}_i^t$ represent the unit motion direction derived from the \textit{Object Trajectory}. \textit{Trajectory Reprojection} then maps a short probe point from the world trajectory to the current frame:
\begin{equation}
(u_i^{t,*},v_i^{t,*})=\Pi_t(\mathbf{p}_i^t+\lambda \hat{\mathbf{v}}_i^t),
\end{equation}
where $\Pi_t(\cdot)$ denotes the camera projection and $\lambda$ is a short probe length. \textit{Motion Trend Generation} uses the reprojected endpoint $(u_i^{t,*},v_i^{t,*})$ to encode the current motion tendency, while \textit{Mask-aware Object Binding} applies the rendering operator $\Gamma(\cdot)$ to place this directional cue on $\Omega_i^t$, yielding $D_i^t=\Gamma(\Omega_i^t,(u_i^t,v_i^t),(u_i^{t,*},v_i^{t,*}))$. Because the arrow is derived from world-coordinate motion rather than raw image displacement, DTV makes true object motion explicit even when camera translation or zoom changes the apparent 2D movement. As a result, DTV is particularly useful for viewpoint-sensitive questions, where the model must distinguish genuine object motion from camera-induced apparent motion.

\paragraph{Dynamic Trace Graph (DTG)}
DTG is the structured textual branch. We partition the video into $K$ temporal windows and summarize each window into DT-Tokens. For object $i$ in window $k$, we first derive three semantic fields from the \textit{Object Trajectory} together with \textit{Camera Behavior}: dynamic cues $\boldsymbol{\delta}_i^k$, trace evolution $\boldsymbol{\tau}_i^k$, and key moments $\boldsymbol{\mu}_i^k$. Here, dynamic cues summarize the object state that is semantically salient for reasoning, such as position, heading, motion direction, speed trend, and camera-relative status. Trace evolution summarizes how these cues change within the window, for example, whether the object keeps moving left, accelerates, or changes its direction. Key moments record sparse but decisive events, such as turns, sudden state changes, or moments when an object becomes most relevant to the query. The Tokenizer then converts these three fields into an object token:
\begin{equation}
\mathbf{o}_i^k=f_{\mathrm{tok}}^{\mathrm{obj}}(\boldsymbol{\delta}_i^k,\boldsymbol{\tau}_i^k,\boldsymbol{\mu}_i^k).
\end{equation}

For an object pair $(i,j)$ in the same window, we similarly derive relation-side dynamic cues $\boldsymbol{\delta}_{ij}^k$, trace evolution $\boldsymbol{\tau}_{ij}^k$, and key moments $\boldsymbol{\mu}_{ij}^k$ from \textit{Relation Evolution}. In this case, dynamic cues describe pairwise states such as distance and bearing, trace evolution records how these relations change over time, and key moments mark events such as closest approach, crossing, or separation. The Tokenizer maps them into a relation token:
\begin{equation}
\mathbf{r}_{ij}^k=f_{\mathrm{tok}}^{\mathrm{rel}}(\boldsymbol{\delta}_{ij}^k,\boldsymbol{\tau}_{ij}^k,\boldsymbol{\mu}_{ij}^k).
\end{equation}
Object tokens summarize per-object dynamics as graph nodes, while relation tokens summarize cross-object dynamics as graph edges. Since both token types are reconstructed from the same window under the same camera context, relation evolution remains consistent with the underlying object trajectories.

Given the object-token set $\mathcal{O}^k=\{\mathbf{o}_i^k\}_i$, the relation-token set $\mathcal{R}^k=\{\mathbf{r}_{ij}^k\}_{i,j}$, and the serialized camera summary $\mathbf{c}^k=\operatorname{ser}(\mathcal{B}^k)$ with $\mathcal{B}^k$ denoting the camera behavior in window $k$, we organize them into a window-level graph and then serialize the graph sequence into DTG:
\begin{equation}
\mathcal{G}^k=f_{\mathrm{graph}}(\mathbf{c}^k,\mathcal{O}^k,\mathcal{R}^k),\quad
\mathcal{G}=\left[\operatorname{ser}(\mathcal{G}^k)\right]_{k=1}^{K}.
\end{equation}
In this form, DTG keeps long-horizon object dynamics and relation changes explicit, while remaining much more compact than frame-wise geometric sequences. This compactness is important
because the final MLLM is relieved of the necessity to reconstruct long-range traces from raw frames, 
but can directly follow dynamic cues, trace evolution, and key moments through the serialized graph.

\subsection{Representation Integration \& Reasoning}
During inference, DynTrace feeds the user query $q$, the DTV collection $\mathcal{D}$, and the serialized DTG $\mathcal{G}$ into the target MLLM:
\begin{equation}
\hat{y}=\mathrm{MLLM}(q,\mathcal{D},\mathcal{G}).
\end{equation}
DTV provides immediate geometry-informed visual priors for accurate motion-source disambiguation, while DTG provides a compact textual scaffold over object and relation evolution across time and interaction states under changing viewpoints and long temporal horizons in complex scenes. For viewpoint-sensitive questions, the model mainly relies on DTV together with the camera behavior explicitly encoded in DTG for correct perspective grounding. For relation and event-timing questions, it follows the dynamic cues, trace evolution, and key moments stored in object and relation tokens in temporal order. Their joint use injects a continuous and verifiable dynamic evidence chain into existing MLLMs, allowing the backbone to focus on multimodal reasoning and response synthesis instead of repeatedly reconstructing long-range dynamics from sparse frame observations alone.

\begin{table*}[t]
\centering
\caption{Comparison on VLM4D and DSI-Bench. The enhanced rows apply DynTrace at inference time. \textbf{Obj-Scn}, \textbf{Obs-Scn}, and \textbf{Obs-Obj} stand for Object-Scene, Observer-Scene, and Observer-Object, respectively. Bold and underlined values denote the best and second-best results, respectively.}
\label{tab:merge_results}
\setlength{\tabcolsep}{5.5pt}
\renewcommand{\arraystretch}{0.9} 

\resizebox{\textwidth}{!}{%
\begin{tabular}{l|*{7}{>{\centering\arraybackslash}p{1.5cm}}}
\toprule
\multirow{2.5}{*}{\textbf{Method}} & \multicolumn{3}{c}{\textbf{VLM4D}} & \multicolumn{4}{c}{\textbf{DSI-Bench}} \\
\cmidrule(l){2-4}\cmidrule(l){5-8}
\noalign{\vskip 1pt}
\rule{0pt}{2.2ex} & \textbf{Avg.} & \cellcolor{orange!20}\textbf{Real} & \cellcolor{yellow!20}\textbf{Synthetic} & \textbf{Avg.} & \cellcolor{orange!20}\textbf{Obj-Scn} & \cellcolor{yellow!20}\textbf{Obs-Scn} & \cellcolor{blue!20}\textbf{Obs-Obj} \\
\midrule

\rowcolor{gray!10}\multicolumn{8}{l}{\textit{Spatial MLLMs}} \\
SpaceR-7B~\cite{spacer} & 47.4 & 49.2 & 41.8 & 54.2 & 71.2 & 38.3 & 51.1 \\
VST-7B-RL~\cite{vst} & 44.7 & 45.1 & 43.1 & 51.5 & 69.4 & 34.8 & 48.4 \\
Spatial-SSRL-7B~\cite{spatial-ssrl} & 52.4 & 52.2 & 53.3 & 51.4 & 66.5 & 37.0 & 50.2 \\
MLLM-4D-8B~\cite{yin2026mllm4d} & \underline{59.1} & \underline{59.3} & 58.4 & 45.2 & 61.3 & 33.4 & 30.9 \\
\midrule

\rowcolor{gray!10}\multicolumn{8}{l}{\textit{Video MLLMs}} \\
LLaVA-OV-1.5-8B~\cite{LLaVA-OneVision-1.5} & 46.3 & 47.8 & 41.3 & 52.7 & 72.2 & 36.6 & 42.2 \\
InternVideo2.5-Chat-8B~\cite{wang2025internvideo} & 49.0 & 50.6 & 44.0 & 49.4 & 64.0 & 35.0 & 49.3 \\
VideoLLaMA3-7B~\cite{damonlpsg2025videollama3} & 49.8 & 55.6 & 32.1 & \underline{55.6} & \textbf{74.8} & 37.5 & 52.9 \\
\midrule

\rowcolor{gray!10}\multicolumn{8}{l}{\textit{Training-free MLLMs}} \\
See\&Trek~\cite{li2025seetrek} & 47.8 & 44.6 & 57.5 & 52.7 & 66.1 & 38.5 & 56.5 \\
GSM~\cite{zhou2025learning} & 48.4 & 48.6 & 47.6 & 54.7 & \underline{73.9} & 36.8 & 51.1 \\
\midrule

\rowcolor{gray!10}\multicolumn{8}{l}{\textit{Ours}} \\
Qwen3-VL-8B-Instruct~\cite{Qwen3-VL} & 59.0 & 58.1 & \underline{61.6} & 53.4 & 66.2 & 39.4 & 58.7 \\
\rowcolor{blue!5}\quad\textbf{+DynTrace} & \textbf{64.2} \small{\textcolor{darkred}{+5.2}} & \textbf{62.9} \small{\textcolor{darkred}{+4.8}} & \textbf{68.3} \small{\textcolor{darkred}{+6.7}} & \textbf{56.4} \small{\textcolor{darkred}{+3.0}} & 67.8 \small{\textcolor{darkred}{+1.6}} & \textbf{43.5} \small{\textcolor{darkred}{+4.1}} & \textbf{63.1} \small{\textcolor{darkred}{+4.4}} \\
InternVL3.5-8B~\cite{wang2025internvl3_5} & 48.6 & 49.5 & 46.1 & 44.5 & 52.1 & 35.0 & 51.6 \\
\rowcolor{blue!5}\quad\textbf{+DynTrace} & 54.7 \small{\textcolor{darkred}{+6.1}} & 53.2 \small{\textcolor{darkred}{+3.7}} & 59.2 \small{\textcolor{darkred}{+13.1}} & 48.7 \small{\textcolor{darkred}{+4.2}} & 53.1 \small{\textcolor{darkred}{+1.0}} & \underline{41.5} \small{\textcolor{darkred}{+6.5}} & \underline{59.0} \small{\textcolor{darkred}{+7.4}} \\
\bottomrule
\end{tabular}%
}
\end{table*}

\section{Experiments}
\label{sec:exp}

\subsection{Experimental Setup}
We evaluate DynTrace on three representative dynamic 4D reasoning benchmarks: Dyn-Bench~\cite{huang2026thinking}, VLM4D~\cite{zhou2025vlm4d}, and DSI-Bench~\cite{zhang2025dsibench}. Dyn-Bench, our main benchmark, systematically tests whether a model can reason with object-centered dynamic evidence consistently across \textit{Inter-Object}, \textit{Object-Scene}, and \textit{Camera-Object} settings. VLM4D contains both real and synthetic videos and emphasizes perspective-aware motion understanding. DSI-Bench focuses on observer-centric dynamic spatial intelligence, where observer and object motion are tightly coupled throughout. For all benchmarks, we follow the official protocols and report overall accuracy.

For backbone comparison, we use Qwen3-VL-8B-Instruct~\cite{Qwen3-VL}, Qwen3-VL-32B-Instruct~\cite{Qwen3-VL}, InternVL3.5-8B~\cite{wang2025internvl3_5}, and InternVL3.5-14B~\cite{wang2025internvl3_5} on Dyn-Bench, and report Qwen3-VL-8B-Instruct and InternVL3.5-8B on VLM4D and DSI-Bench. We compare each backbone with its DynTrace-enhanced version. We also include open-source competitors from three families, Spatial MLLMs, Video MLLMs, and training-free MLLMs, to assess our method more broadly. Unless otherwise stated, we uniformly sample 16 frames from each video. Rather than feeding original video frames to the target MLLM, the DynTrace-enhanced version feeds DTV together with the serialized DTG and the query. Thus, the target MLLM receives both geometry-informed visual priors and continuous structured dynamic evidence, designed to mitigate \textit{Dynamic Source Confusion} and \textit{Dynamic Trace Loss}.

\subsection{Quantitative Results}
\textbf{Overall.} 
DynTrace consistently improves all evaluated backbones across the three benchmarks. On Dyn-Bench, the four backbone variants gain 4.3\% to 5.0\% on average. On VLM4D, Qwen3-VL-8B and InternVL3.5-8B improve by 5.2\% and 6.1\%, and on DSI-Bench by 3.0\% and 4.2\%, respectively. This shows that the benefit is stable across benchmark styles rather than tied to a specific setting. This robustness matters because the benchmarks stress different aspects of dynamic 4D reasoning, including trace preservation, viewpoint-aware motion interpretation, and observer-centric spatial understanding. The consistent gains indicate DynTrace addresses a shared dynamic object evidence bottleneck rather than fitting one dataset style.

\textbf{Dyn-Bench.} Dyn-Bench serves as our primary benchmark, designed for dynamic-object-centric reasoning across \textit{Inter-Object}, \textit{Object-Scene}, and \textit{Camera-Object} configurations. This framework directly instantiates our central premise: accurate MLLM reasoning requires continuous, interpretable dynamic object evidence, especially under complex viewpoint variations.
As shown in Tab.~\ref{tab:dynbench_results}, DynTrace improves all four backbones and achieves the best overall result. The gain is not merely from scaling the backbone. Qwen3-VL-8B + DynTrace surpasses the raw Qwen3-VL-32B baseline, showing that explicit dynamic evidence can be more valuable than a larger backbone. The improvement is also meaningful relative to other open-source baselines, including Spatial MLLMs and training-free MLLMs, suggesting that generic spatial priors or coarse inference-time cues remain insufficient when tasks depend on continuous dynamic reasoning. The fine-grained results further verify the roles of DTG and DTV. Dyn-Bench defines \textit{Inter-Object} around motion interactions and spatial relations among multiple dynamic objects, such as approach, occlusion, and overtaking. This is where relation-side DT-Token and DTG help most, because they preserve how relative distance, bearing, and interaction status evolve over time rather than exposing only sparse end states. Camera-aware categories also improve clearly, consistent with DTV separating true object motion from camera-induced apparent motion and making viewpoint-sensitive dynamic cues more robust.
Taken together, Dyn-Bench provides the clearest validation of our design. DTG preserves trajectory continuity and cross-object relation evolution, while DTV corrects motion interpretation under camera movement. Their combination gives the model an explicit dynamic evidence chain, which is why Dyn-Bench offers the strongest and most direct support for DynTrace.

\begin{figure*}[t]
    \centering
    \includegraphics[width=1\textwidth]{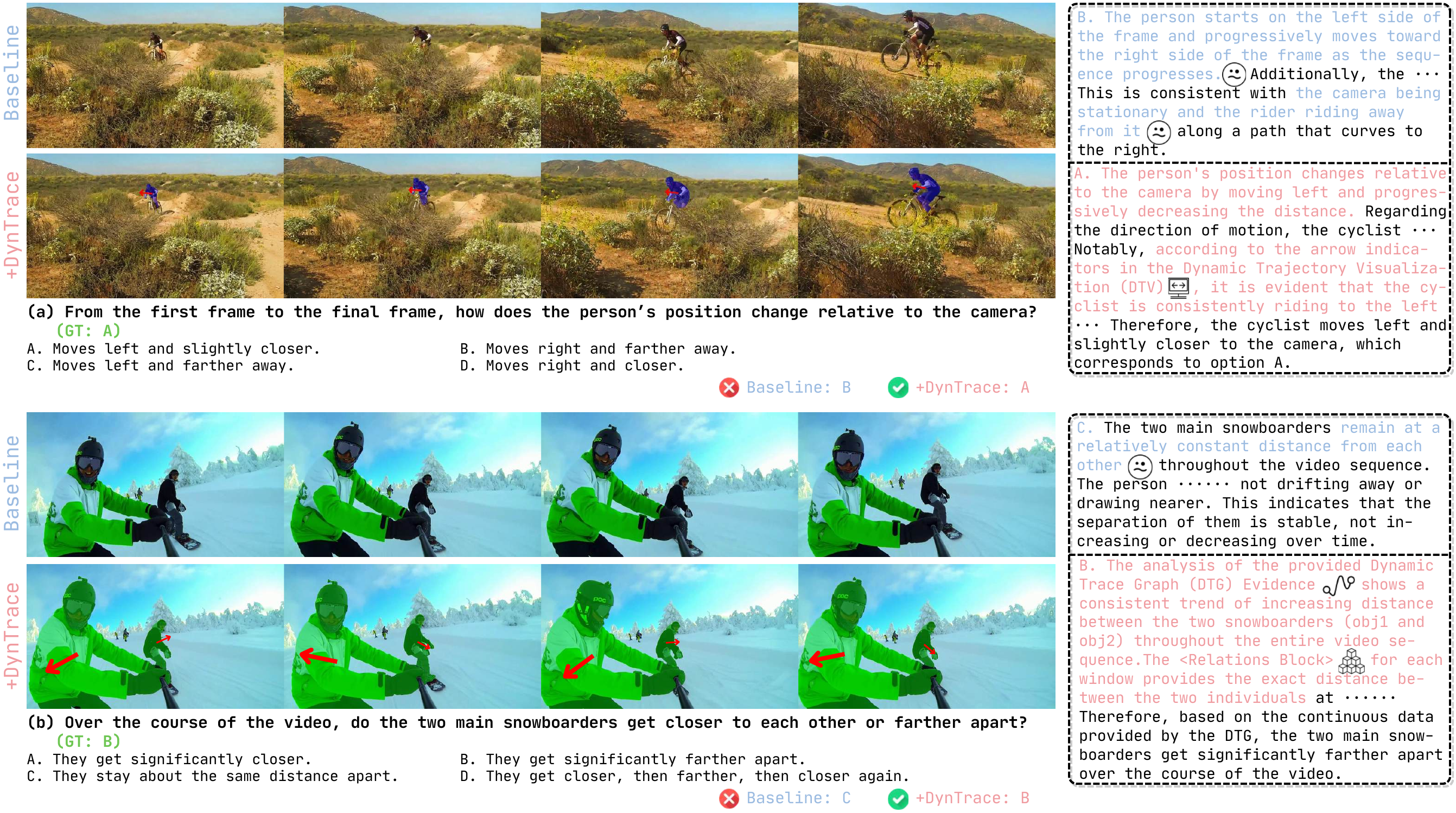}
    \caption{Qualitative results of DynTrace on two representative cases. The figure shows how DynTrace changes the evidence path for reasoning: DTV corrects ambiguous camera-relative motion in case (a), while DTG preserves long-range relation traces in case (b), leading to correct predictions in both cases. }
    \label{fig:qualitative_cases}
\end{figure*}

\textbf{VLM4D and DSI-Bench.} 
Tab.~\ref{tab:merge_results} presents results on VLM4D and DSI-Bench, which complement Dyn-Bench. VLM4D probes MLLM spatio-temporal awareness, focusing on motion reasoning under perspective shifts and temporal continuity. This aligns with DynTrace, which addresses the discrepancy between raw 2D appearance variation and robust dynamic object evidence. We observe consistent gains across both backbones on VLM4D, especially on the synthetic split. This success reflects synthetic data characteristics: high visual quality, stable camera motion, and distinct foreground objects with steady trajectories. Under these conditions, DTV extracts more precise dynamic cues while DTG preserves a more stable trace, facilitating more effective MLLM reasoning.

DSI-Bench evaluates dynamic spatial intelligence by decoupling observer motion from object motion. The most pronounced improvement occurs on \textit{Obs-Obj}, which necessitates tracking the evolution of object dynamics relative to a moving observer under continuously changing viewpoints throughout the sequence. Within this setting, our two components prove highly complementary: DTV mitigates observer-induced motion ambiguity at the visual level, while DTG maintains the observer-object relational trace across time in an explicitly structured and queryable form. These gains indicate that DynTrace enhances not only scene-level motion interpretation but also observer-centric spatial reasoning that depends on stable relational grounding. Collectively, VLM4D and DSI-Bench demonstrate that DynTrace is benchmark-agnostic, consistently converting complex dynamic scenes into geometry-informed visual priors and temporally structured evidence, both of which are highly amenable to reliable MLLM inference.

\begin{table}[h]
    \centering
    \caption{Ablation studies on Dyn-Bench. 
    The upper block evaluates the contributions of DTV and DTG, and the lower block evaluates three components in DT-Token.
    }
    \label{tab:comprehensive_ablation_compact}
    \resizebox{\linewidth}{!}{%
    \begin{tabular}{l|c|ccc}
    \toprule
    \textbf{Variant} & \textbf{Avg.} & \cellcolor{orange!20} \textbf{\scriptsize Inter-Object} & \cellcolor{yellow!20} \textbf{\scriptsize Object-Scene} & 
    \cellcolor{blue!15} \textbf{\scriptsize Camera-Object} \\
    \midrule
    \textbf{Qwen3-VL-8B}       & 60.8 & 56.7 & 74.2 & 54.7 \\
    + DTG                      & 63.0 & 61.8 & 76.3 & 55.0 \\
    + DTV                      & 61.7 & 59.3 & 74.5 & 54.8 \\
    + DTV \& DTG               & \textbf{65.8} & 63.6 & 79.6 & 58.0 \\
    \midrule
    \textbf{InternVL3.5-8B}    & 53.2 & 50.8 & 66.8 & 45.9 \\
    + Cues                     & 56.8 & 55.5 & 70.6 & 48.4 \\
    + Cues \& Trace            & 57.6 & 56.6 & 70.8 & 50.1 \\
    + Cues \& Trace \& Moments & \textbf{58.0} & 56.8 & 72.3 & 50.3 \\
    \bottomrule
    \end{tabular}%
    }
\end{table}

\subsection{Qualitative Results}
As shown in Fig.~\ref{fig:qualitative_cases}, we present two representative cases to analyze why DynTrace improves dynamic 4D reasoning. More results can be found in the appendix. Case (a) is a camera-relative motion scenario in which a cyclist moves across the scene while the viewpoint also changes. Here, DTV plays the primary role by providing geometry-informed visual priors that clarify the true motion direction on the image plane, while DTG still contributes temporal continuity and camera-aware context. Case (b) is a long-horizon interaction scenario involving two snowboarders whose distance gradually changes over time. In this case, DTG plays the primary role by preserving a consistent relation trace across temporal windows, while DTV still helps maintain locally grounded motion evidence under the shared viewpoint. Overall, the qualitative results show that DynTrace improves reasoning not through a single branch alone, but by combining DTV and DTG to make dynamic object evidence more continuous and more interpretable for the MLLM.

\subsection{Further Empirical Study}

\textbf{Ablation of DTV and DTG.} Tab.~\ref{tab:comprehensive_ablation_compact} presents the ablation study of DTV and DTG using Qwen3-VL-8B in the upper block. DTG alone raises the average score from 60.8\% to 63.0\%, and DTV alone improves it to 61.7\%, showing that the former mainly preserves trace continuity while the latter mainly resolves viewpoint-induced motion ambiguity. Their combination further reaches 65.8\% and performs best across all three groups. This also indicates that the two branches are complementary rather than redundant: DTG stabilizes long-range reasoning, while DTV makes the visual motion evidence easier to interpret under camera change.

\textbf{Ablation of DT-Token components.} Tab.~\ref{tab:comprehensive_ablation_compact} presents the ablation study for DT-Token using InternVL3.5-8B in the lower block. \textit{Cues} provide the main gain, improving the average score from 53.2\% to 56.8\%. Adding \textit{Trace} further lifts it to 57.6\% and improves \textit{Inter-Object} from 55.5\% to 56.6\%, while adding \textit{Moments} reaches the best overall result of 58.0\%. This progression shows that cues anchor the current state, trace preserves its evolution, and moments retain decisive events. In other words, DT-Token is most effective when it preserves both stable dynamic states and the sparse temporal anchors that mark important changes.

\begin{figure}[h]
\centering
\includegraphics[width=\columnwidth]{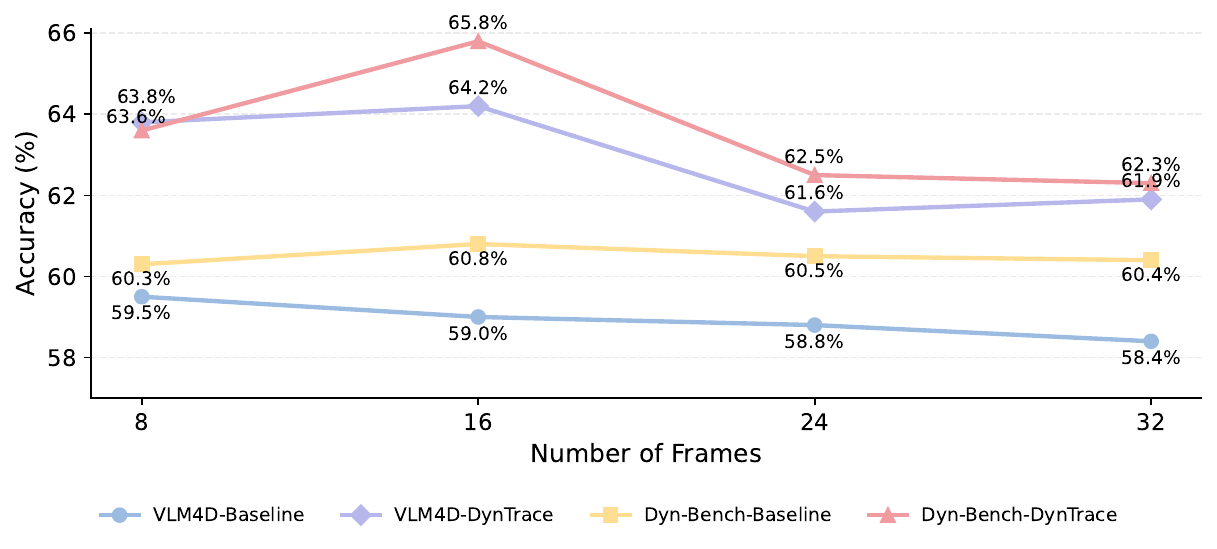}
\caption{Effect of frame sampling on Dyn-Bench and VLM4D.}
\label{fig:frame_sampling}
\end{figure}

\textbf{Frame Sampling.} Fig.~\ref{fig:frame_sampling} evaluates the impact of frame density on Qwen3-VL-8B, which directly pertains to \textit{Dynamic Trace Loss}. DynTrace consistently outperforms the baseline across all tested frame counts on Dyn-Bench and VLM4D, with the best results at 16 frames, reaching 65.8\% and 64.2\%, respectively. Moderate sampling works best: too few frames weaken the temporal evidence chain, while denser sampling becomes redundant once DTG already preserves the dynamic trace.

\begin{figure}[h]
\centering
\includegraphics[width=\columnwidth]{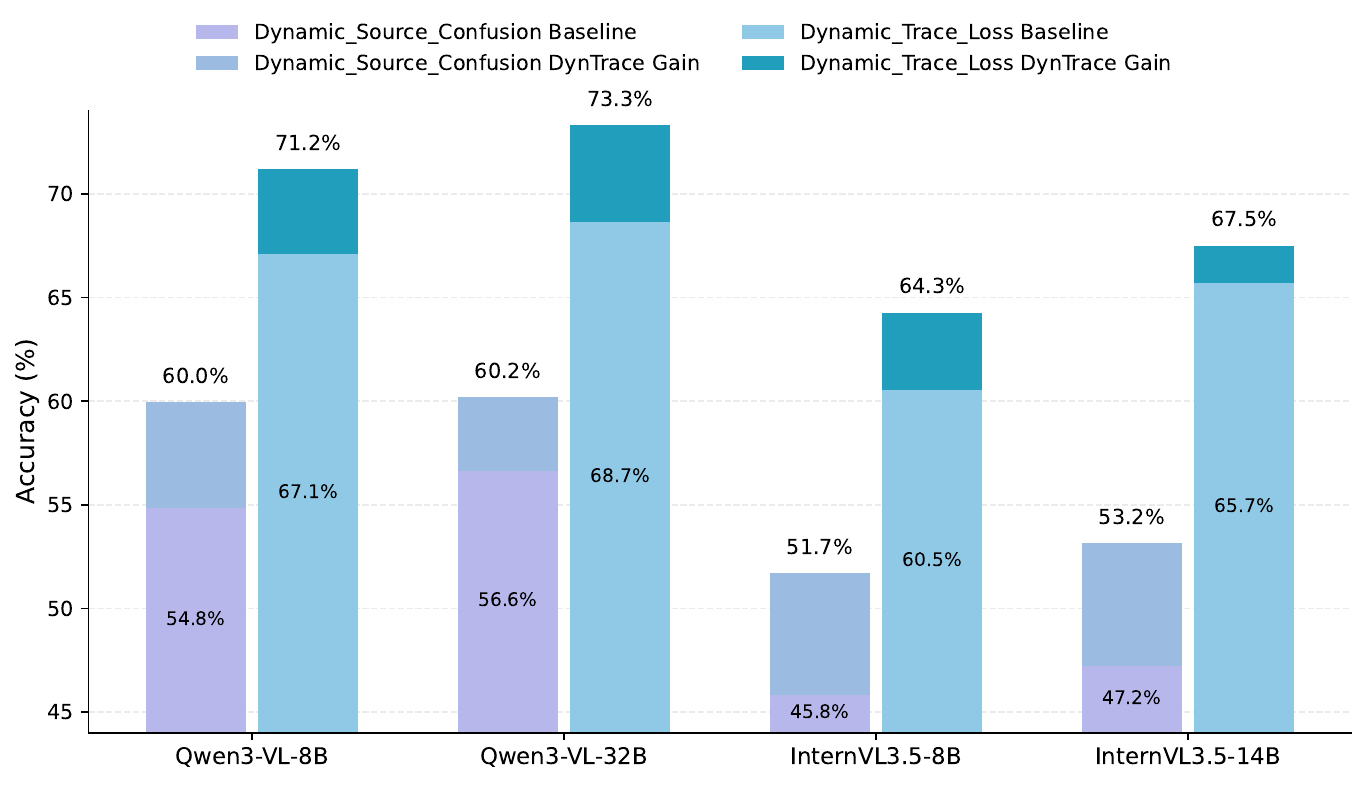}
\caption{Performance on the Dynamic Source Confusion and Dynamic Trace Loss subsets of Dyn-Bench.}
\label{fig:failure_mode_subsets}
\end{figure}

\textbf{Failure-mode Subsets Analysis.} Fig.~\ref{fig:failure_mode_subsets} evaluates the subsets of Dyn-Bench most directly related to the two bottlenecks addressed in this work across four backbones. We define 15 rules and use Qwen3-VL-32B to classify the questions into \textit{Dynamic Source Confusion} and \textit{Dynamic Trace Loss} subsets, accounting for 48.8\% and 43.7\% of Dyn-Bench. Detailed rules are provided in the appendix. DynTrace yields consistent gains on both subsets for all four models, supporting the claim that DTV disentangles apparent motion from true motion while DTG keeps dynamic evidence continuous.

\vspace{-1em}

\begin{figure}[h]
\centering
\includegraphics[width=0.55\columnwidth]{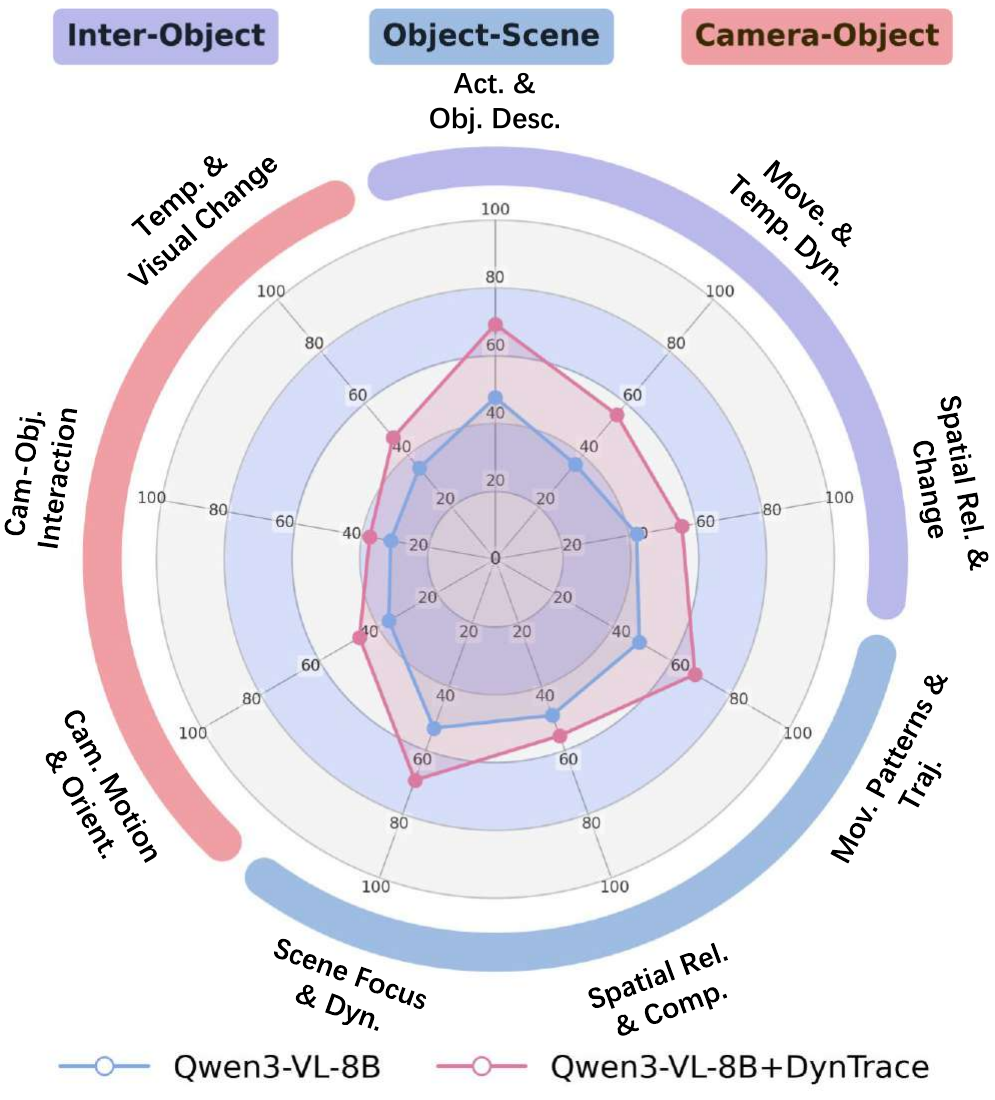}
\caption{Performance comparison on Dyn-Bench-200.}
\label{fig:hardest_radar}
\end{figure}

\textbf{Hardest Subset Analysis.} Fig.~\ref{fig:hardest_radar} compares Qwen3-VL-8B performance on Dyn-Bench-200, the 200 most challenging scenes in Dyn-Bench. DynTrace improves all nine categories, including both motion-centric and camera-related ones. This shows that the method remains effective even in difficult scenarios requiring simultaneous trace preservation and motion-source disentanglement.

\vspace{-1em}

\section{Conclusion}
\label{sec:conclusion}
DynTrace addresses two evidence deficits causing MLLMs to fail in dynamic 4D spatio-temporal reasoning. The first is \textit{Dynamic Source Confusion}, where models conflate genuine object dynamics with camera-induced apparent motion. The second is \textit{Dynamic Trace Loss}, where sparse frame sampling fragments object trajectories, breaking the dynamic evidence chain.
DTV resolves the first by reprojecting world trajectories onto the image plane, yielding geometry-informed visual priors to rectify apparent motion bias.
Complementarily, DTG resolves the second by organizing dynamic cues, trace evolution, and key moments into a queryable graph, preserving long-horizon dynamic continuity.
Across Dyn-Bench, VLM4D, and DSI-Bench, DynTrace consistently improves open-source MLLMs, establishing that tracking dynamic object evidence is important for robust 4D spatio-temporal reasoning.

{
    \small
    \bibliographystyle{ieeenat_fullname}
    \bibliography{main}
}

\clearpage
\appendix
This supplementary material complements the main paper \textit{"DynTrace: Tracking Dynamic Object Evidence for 4D Spatio-Temporal Reasoning in MLLMs"} by providing further implementation details, empirical analyses, and visualizations, organized as follows:
\begin{itemize}
    \item \textbf{Section ~\ref{app:methods}} details the implementation of DynTrace, including the inference pipeline, Dynamic Objects Extraction, Spatio-Temporal Dynamics Encoding, and Representation Integration \& Reasoning.
    \item \textbf{Section ~\ref{app:exp}} introduces the three evaluation benchmarks, reporting additional quantitative analyses, robustness studies, and limitation cases.
    \item \textbf{Section ~\ref{app:additonal}} provides additional visualizations and prompt templates used for query-guided entity discovery, multi-frame box verification, final MLLM reasoning, and failure-mode subset selection.
\end{itemize}

\section{Methods}
\label{app:methods}

\subsection{Dynamic Objects Extraction}
\label{app:dynamic_object_extraction}
\paragraph{Query-guided entity discovery.}
The first stage begins by sending the video and the query to Qwen3-VL-8B to identify the primary dynamic entities relevant to the question. The corresponding prompt template is summarized in Section~\ref{app:prompt_templates}. The resulting entity set is denoted by
\begin{equation}
\mathcal{E}^{(0)} = \{e_1, e_2, \ldots, e_N\} = \mathrm{MLLM}(\mathcal{V}, q).
\end{equation}
Each $e_i$ is a textual description of one query-relevant dynamic object instance and is later used as a text prompt for SAM3.

\paragraph{Motion-region box extraction.}
For consecutive frames, WAFT estimates dense optical flow $\mathbf{F}_t$. To suppress camera-dominated motion, an affine transform is fitted to sampled flow correspondences:
\begin{equation}
\begin{bmatrix}
\hat{x} \\
\hat{y}
\end{bmatrix}
=
\mathbf{A}_t
\begin{bmatrix}
 x \\
 y \\
 1
\end{bmatrix}.
\end{equation}
This produces the background flow field
\begin{equation}
\hat{\mathbf{F}}_{t}^{\mathrm{bg}}(x,y)=
\begin{bmatrix}
\hat{x}-x \\
\hat{y}-y
\end{bmatrix},
\end{equation}
and the residual flow
\begin{equation}
\mathbf{F}_{t}^{\mathrm{res}} = \mathbf{F}_t - \hat{\mathbf{F}}_{t}^{\mathrm{bg}}.
\end{equation}
The residual magnitude map is then thresholded as
\begin{equation}
\Omega_t(x,y)=\mathbf{1}\big(\lVert \mathbf{F}_{t}^{\mathrm{res}}(x,y) \rVert_2 > \tau_f\big),
\end{equation}
followed by erosion and morphological opening. The remaining motion regions are clustered into object-level proposals. Each cluster is expanded back to a motion box:
\begin{equation}
\mathcal{B}_t = \{b_{t,m}\}_{m=1}^{M_t},
\qquad
b_{t,m} = [x_{t,m}, y_{t,m}, w_{t,m}, h_{t,m}].
\end{equation}

\paragraph{Box verification and point sampling.}
The corresponding multi-frame verification prompt template is summarized in Section~\ref{app:prompt_templates}. Here we focus on the resulting verification process and feed the frames annotated with box IDs back into Qwen3-VL-8B for multi-frame verification:
\begin{equation}
(\mathcal{E}, \mathcal{B}^{\mathrm{val}}) = \mathrm{Qwen3\mbox{-}VL}(\mathcal{V}, q, \mathcal{E}^{(0)}, \{\mathcal{B}_t\}),
\end{equation}
where $\mathcal{B}^{\mathrm{val}}_t$ keeps only the valid box IDs on frame $t$. Positive points are then sampled only from the retained boxes:
\begin{equation}
\mathcal{S}_t = \{\mathbf{s}_{t,m,n}\mid b_{t,m} \in \mathcal{B}^{\mathrm{val}}_t\}.
\end{equation}
The final point set is obtained by merging all retained samples,
\begin{equation}
\mathcal{S}=\bigcup_{t=1}^{T-1}\mathcal{S}_t.
\end{equation}

\paragraph{Text-first and point-refined SAM3.}
The retained entity descriptions are first used to initialize SAM3 tracking,
\begin{equation}
\mathcal{M}^{(0)} = \mathrm{SAM3}_{\mathrm{text}}(\mathcal{V}, \mathcal{E}),
\end{equation}
and the retained point prompts are then used for refinement,
\begin{equation}
\mathcal{M} = \mathrm{SAM3}_{\mathrm{point}}(\mathcal{M}^{(0)}, \mathcal{S}).
\end{equation}

\subsection{Spatio-Temporal Dynamics Encoding}
\label{app:spatio_temporal_encoding}
\paragraph{Geometry-grounded dynamic evidence.}
Given the dynamic masks $\mathcal{M}$, we use DA3 to estimate depth, intrinsics, and camera pose for each frame. For object $i$ at frame $t$, we denote the centroid of mask $M_i^t$ by $(u_i^t,v_i^t)$, the median depth inside the mask by $z_i^t$, the intrinsic matrix by $\mathbf{K}_t$, and the camera-to-world transform by $\mathbf{T}_{c\rightarrow w}^{t}$. We then recover the object position in world coordinates as
\begin{equation}
\mathbf{p}_i^t = \mathbf{T}_{c\rightarrow w}^{t}\left(z_i^t\mathbf{K}_t^{-1}[u_i^t, v_i^t, 1]^\top\right).
\end{equation}
Across frames, $\mathcal{T}_i=\{\mathbf{p}_i^t\}_{t=1}^{T}$ forms the \textit{Object Trajectory}. For each object pair $(i,j)$, we compute the metric distance $d_{ij}^t=\lVert \mathbf{p}_i^t-\mathbf{p}_j^t\rVert_2$ and the relative bearing $b_{ij}^t=f_{\mathrm{bear}}(\mathbf{p}_j^t-\mathbf{p}_i^t)$, and use their temporal sequence to describe \textit{Relation Evolution}. We further analyze the camera poses to summarize \textit{Camera Behavior}, including the dominant translation trend, viewpoint stability, and scale change. These three streams provide the shared geometric basis for both DTV and DTG.

\paragraph{DTV generation.}
We estimate a unit motion direction $\hat{\mathbf{v}}_i^t$ from the local trajectory of object $i$, reproject a short probe point to the image plane by $(u_i^{t,*}, v_i^{t,*}) = \Pi_t(\mathbf{p}_i^t + \lambda \hat{\mathbf{v}}_i^t)$, and adaptively enlarge the probe length $\lambda$ until the projected displacement is visually distinguishable. We then bind the projected endpoint to the current mask support $\Omega_i^t$ and render a directional cue:
\begin{equation}
D_i^t = \Gamma(\Omega_i^t, (u_i^t, v_i^t), (u_i^{t,*}, v_i^{t,*})).
\end{equation}
Collecting all rendered cues gives the Dynamic Trajectory Visualization $\mathcal{D}=\{D_i^t\mid i=1,\ldots,N_t,\; t=1,\ldots,T\}$. Compared with raw image-plane motion, this representation makes object movement more robust to camera-induced drift.

\paragraph{DT-Token and DTG construction.}
We partition the video into temporal windows and summarize each window into structured evidence. For object $i$ in window $k$, we extract dynamic cues from its end-of-window state, including position $\mathbf{p}_{i,\mathrm{end}}^k$, speed $s_{i,\mathrm{end}}^k$, heading $h_{i,\mathrm{end}}^k$, and camera-relative state $\gamma_{i,\mathrm{cam}}^k$. We encode trace evolution by summarizing how the object changes within the window, such as its dominant direction and speed profile, and we record key moments such as turning, stopping, or abrupt state change. We then tokenize these three parts into an object token:
\begin{equation}
\mathbf{o}_i^k = \operatorname{Tok}(\boldsymbol{\delta}_i^k, \boldsymbol{\tau}_i^k, \boldsymbol{\mu}_i^k).
\end{equation}
In this token, $\boldsymbol{\delta}_i^k$ stores object-side dynamic cues, $\boldsymbol{\tau}_i^k$ stores object-side trace evolution, and $\boldsymbol{\mu}_i^k$ stores object-side key moments.

For each valid object pair $(i,j)$ in the same window, we summarize relation-side dynamic cues from the end-of-window distance $d_{ij,\mathrm{end}}^k$ and bearing $b_{ij,\mathrm{end}}^k$, relation-side trace evolution from how distance and bearing evolve over time, and relation-side key moments from decisive interaction events such as closest approach, crossing, or clear separation. We tokenize them into a relation token:
\begin{equation}
\mathbf{r}_{ij}^k = \operatorname{Tok}(\boldsymbol{\delta}_{ij}^k, \boldsymbol{\tau}_{ij}^k, \boldsymbol{\mu}_{ij}^k).
\end{equation}
For the relation token, $\boldsymbol{\delta}_{ij}^k$, $\boldsymbol{\tau}_{ij}^k$, and $\boldsymbol{\mu}_{ij}^k$ respectively store relation-side dynamic cues, trace evolution, and key moments.

We then organize each temporal window as a graph rather than a flat token list. Specifically, we use the object tokens as graph nodes and instantiate a relation edge only when the corresponding objects co-exist and form a valid pair in that window. The camera summary $\mathbf{c}^k$ acts as shared global context for all nodes and edges. Let $\mathcal{O}^k=\{\mathbf{o}_i^k\}_i$ denote the node set and $\mathcal{R}^k=\{\mathbf{r}_{ij}^k\}_{(i,j)\in\mathcal{P}^k}$ denote the edge set over valid pairs $\mathcal{P}^k$. We organize the window-level graph as
\begin{equation}
\mathcal{G}^k = (\mathbf{c}^k, \mathcal{O}^k, \mathcal{R}^k),
\end{equation}
where $\mathbf{c}^k$ is placed before the node and edge tokens during serialization, so each window keeps a consistent global-to-local layout. We then serialize all window graphs in temporal order to obtain the final DTG:
\begin{equation}
\mathcal{G} = [\operatorname{ser}(\mathcal{G}^k)]_{k=1}^{K}.
\end{equation}
In practice, we serialize each window by placing the camera summary first, followed by object tokens ordered by object identity and relation tokens ordered by object-pair identity, so that the MLLM receives a stable textual layout across windows. This organization preserves both per-object dynamics and cross-object interactions while remaining substantially more compact than frame-wise geometric traces.
\subsection{Representation Integration \& Reasoning}
\label{app:integration_reasoning}

\begin{figure}[h]
    \centering
    \includegraphics[width=0.8\columnwidth]{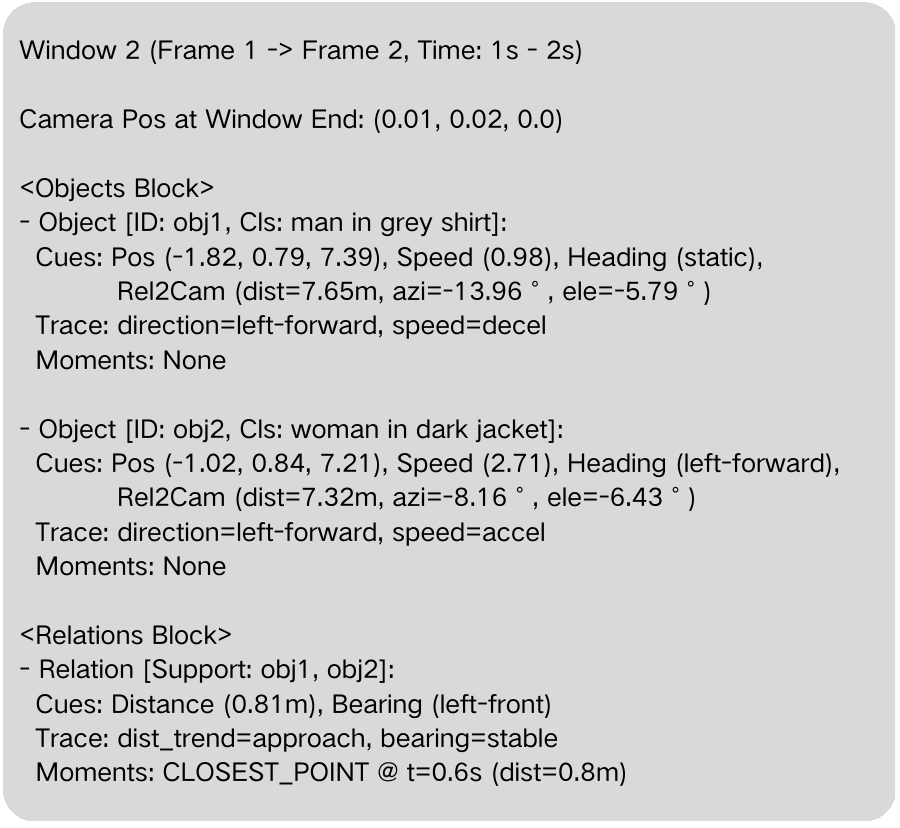}
    \caption{Example of serialized DTG.}
    \label{fig:appendix_dtg_example}
\end{figure}
The final stage serializes the window-level graph sequence $\mathcal{G}$ into textual DTG evidence $\widetilde{\mathcal{G}}$, which contains a global camera block and window-aligned object and relation blocks. An example of this representation is shown in Fig.~\ref{fig:appendix_dtg_example}. The final reasoning input $\mathcal{P} = [q; \mathcal{D}; \widetilde{\mathcal{G}}]$ is then formed by combining the query $q$, DTV $\mathcal{D}$, and $\widetilde{\mathcal{G}}$. The target MLLM outputs the final answer by:
\begin{equation}
\hat{y} = \mathrm{MLLM}(\mathcal{P}).
\end{equation}
The complete final reasoning prompt template is summarized in Section~\ref{app:prompt_templates}, where DTV provides visual grounding and DTG provides structured dynamic evidence.

\section{Experiments}
\label{app:exp}

\subsection{Benchmark Overview}
\textit{Dyn-Bench}~\cite{huang2026thinking} serves as our primary benchmark for dynamic 4D reasoning and contains 1,000 videos, 7,000 VQA pairs, and 3,000 dynamic grounding annotations. It organizes the VQA task into nine categories under three groups: \textit{Inter-Object}, including \textit{Activity \& Object Description}, \textit{Movement \& Temporal Dynamics}, and \textit{Spatial Relationships \& Change}; \textit{Object-Scene}, including \textit{Movement Patterns \& Trajectories}, \textit{Spatial Relationships \& Composition}, and  \textit{Scene Focus \& Dynamics};  and \textit{Camera-Object},

\clearpage

\begin{table*}[t]
\centering

\caption{Detailed results on VLM4D. Bold and underlined values denote the best and second-best results, respectively.}
\label{tab:vlm4d_results}
{
\setlength{\tabcolsep}{12pt}
\renewcommand{\arraystretch}{0.9}
\begin{tabular}{@{\hspace{4pt}}l|c|cc|cc}
\toprule
\multirow{2}{*}{\textbf{Method}} &
\multirow{2}{*}{\textbf{Avg}} &
\multicolumn{2}{c|}{\cellcolor{blue!15}\textbf{Real}} &
\multicolumn{2}{c}{\cellcolor{yellow!25}\textbf{Synthetic}} \\
& & \textbf{Ego-centric} & \textbf{Exo-centric} & \textbf{Directional} & \textbf{FP} \\
\midrule

\rowcolor{gray!10}[4pt][4pt]\multicolumn{6}{@{\hspace{4pt}}l@{\hspace{4pt}}}{\textit{Spatial MLLMs}} \\
SpaceR-7B~\cite{spacer} & 47.4 & 52.3 & 47.7 & 39.0 & 66.7 \\
VST-7B-RL~\cite{vst} & 44.7 & 49.4 & 43.1 & 42.8 & 46.7 \\
Spatial-SSRL-7B~\cite{spatial-ssrl} & 52.4 & 56.6 & 50.0 & 54.1 & 45.5 \\
SpatialReasoner~\cite{ma2025spatialreasoner} & 42.5 & 51.0 & 38.8 & 44.9 & 9.1 \\
SpatialThinker-7B~\cite{batra2025spatialthinker} & 51.4 & 53.7 & 48.1 & 53.8 & 75.6 \\
MLLM-4D-8B~\cite{yin2026mllm4d} & 59.1 & 53.5 & \underline{62.2} & 60.5 & 40.0 \\
LLaVA-ST-7B~\cite{li2025llavast} & 37.1 & 41.2 & 35.1 & 41.0 & 2.2 \\

\midrule
\rowcolor{gray!10}[4pt][4pt]\multicolumn{6}{@{\hspace{4pt}}l@{\hspace{4pt}}}{\textit{Video MLLMs}} \\
LLaVA-OV-1.5-8B~\cite{LLaVA-OneVision-1.5} & 46.3 & 47.2 & 48.2 & 36.5 & \textbf{84.4} \\
Videorefer-7B~\cite{yuan2025videorefer} & \underline{59.2} & \textbf{61.7} & 58.9 & 57.0 & 60.0 \\
InternVideo2.5-Chat-8B~\cite{wang2025internvideo} & 49.0 & 53.5 & 49.2 & 45.8 & 28.9 \\
VideoLLaMA3-7B~\cite{damonlpsg2025videollama3} & 49.8 & 61.0 & 52.9 & 26.8 & \underline{80.0} \\

\midrule
\rowcolor{gray!10}[4pt][4pt]\multicolumn{6}{@{\hspace{4pt}}l@{\hspace{4pt}}}{\textit{Training-free MLLMs}} \\
See\&Trek~\cite{li2025seetrek} & 47.8 & 48.6 & 42.7 & 57.5 & 57.8 \\
GSM~\cite{zhou2025learning} & 48.4 & 53.9 & 46.0 & 45.0 & 71.1 \\

\midrule
\rowcolor{gray!10}[4pt][4pt]\multicolumn{6}{@{\hspace{4pt}}l@{\hspace{4pt}}}{\textit{Ours}} \\
Qwen3-VL-8B-Instruct~\cite{Qwen3-VL} & 59.0 & 60.1 & 57.2 & \underline{62.3} & 54.6 \\
\rowcolor{blue!5}\multicolumn{1}{@{\hspace{4pt}}>{\columncolor{blue!5}[4pt][12pt]}l|}{\quad \textbf{+DynTrace}} & \textbf{64.2} \small{\textcolor{darkred}{+5.2}} & \underline{61.4} \small{\textcolor{darkred}{+1.3}} & \textbf{63.7} \small{\textcolor{darkred}{+6.5}} & \textbf{69.1} \small{\textcolor{darkred}{+6.8}} & 61.4 \small{\textcolor{darkred}{+6.8}} \\
InternVL3.5-8B~\cite{wang2025internvl3_5} & 48.6 & 47.9 & 50.2 & 46.4 & 43.2 \\
\rowcolor{blue!5}\multicolumn{1}{@{\hspace{4pt}}>{\columncolor{blue!5}[4pt][12pt]}l|}{\quad \textbf{+DynTrace}} & 54.7 \small{\textcolor{darkred}{+6.1}} & 51.4 \small{\textcolor{darkred}{+3.5}} & 53.9 \small{\textcolor{darkred}{+3.7}} & 61.0 \small{\textcolor{darkred}{+14.6}} & 43.2 \small{\textcolor{darkred}{+0.0}} \\

\bottomrule
\end{tabular}%
}

\vspace{0.3cm}

\caption{Detailed results on DSI-Bench. Bold and underlined values denote the best and second-best results, respectively.}
\label{tab:dsibench_results}
{
\setlength{\tabcolsep}{4pt}
\renewcommand{\arraystretch}{1.1}

\begin{tabular}{l|c|cc|cc|cc}
\toprule
\multirow{2}{*}{\textbf{Method}} &
\multirow{2}{*}{\textbf{Avg}} &
\multicolumn{2}{c|}{\cellcolor{orange!25}\textbf{Object-Scene}} &
\multicolumn{2}{c|}{\cellcolor{yellow!25}\textbf{Observer-Scene}} &
\multicolumn{2}{c}{\cellcolor{blue!15}\textbf{Observer-Object}} \\
& & \textbf{Fixed-Obs.} & \textbf{Dyn-Obs.} & \textbf{Static-Sce.} & \textbf{Dyn-Sce.} & \textbf{Distance} & \textbf{Orientation} \\
\midrule

\rowcolor{gray!10}\multicolumn{8}{l}{\textit{Spatial MLLMs}} \\
SpaceR-7B~\cite{spacer} & 54.2 & 70.8 & 71.3 & 37.0 & 38.8 & 65.2 & 28.2 \\
VST-7B-RL~\cite{vst} & 51.5 & 69.7 & 69.2 & 31.7 & 36.1 & 66.7 & 18.8 \\
Spatial-SSRL-7B~\cite{spatial-ssrl} & 51.4 & 61.6 & 68.0 & 41.3 & 35.2 & 60.9 & 32.9 \\
SpatialReasoner~\cite{ma2025spatialreasoner} & 51.0 & 66.5 & 69.6 & 33.0 & 34.8 & 59.4 & 30.6 \\
SpatialThinker-7B~\cite{batra2025spatialthinker} & 52.1 & 71.9 & 71.5 & 33.0 & 34.4 & 63.0 & 23.5 \\
MLLM-4D-8B~\cite{yin2026mllm4d} & 45.2 & 54.1 & 63.6 & 24.8 & 37.0 & 29.7 & 32.9 \\
LLaVA-ST-7B~\cite{li2025llavast} & 38.2 & 48.1 & 46.6 & 29.1 & 25.9 & 55.8 & 35.3 \\

\midrule
\rowcolor{gray!10}\multicolumn{8}{l}{\textit{Video MLLMs}} \\
LLaVA-OV-1.5-8B~\cite{LLaVA-OneVision-1.5} & 52.7 & 73.0 & 72.0 & 31.7 & 38.6 & 43.5 & \textbf{40.0} \\
Videorefer-7B~\cite{yuan2025videorefer} & 52.7 & 66.5 & 68.9 & 33.0 & 39.0 & 61.6 & \textbf{40.0} \\
InternVideo2.5-Chat-8B~\cite{wang2025internvideo} & 49.4 & 66.5 & 63.2 & 33.9 & 35.5 & 61.6 & 29.4 \\
VideoLLaMA3-7B~\cite{damonlpsg2025videollama3} & \underline{55.6} & \underline{74.6} & \textbf{74.9} & 30.0 & \textbf{40.6} & 65.2 & 32.9 \\

\midrule
\rowcolor{gray!10}\multicolumn{8}{l}{\textit{Training-free MLLMs}} \\
See\&Trek~\cite{li2025seetrek} & 52.7 & 60.5 & 67.9 & 35.7 & 39.7 & 69.6 & 35.3 \\
GSM~\cite{zhou2025learning} & 54.7 & \textbf{75.3} & \underline{73.5} & 35.5 & 37.2 & 65.2 & 28.2 \\

\midrule
\rowcolor{gray!10}\multicolumn{8}{l}{\textit{Ours}} \\
Qwen3-VL-8B-Instruct~\cite{Qwen3-VL} & 53.4 & 58.8 & 68.6 & 39.1 & 39.5 & 72.5 & \underline{36.5} \\
\rowcolor{blue!5}\quad \textbf{+DynTrace} & \textbf{56.4} \small{\textcolor{darkred}{+3.0}} & 62.6 \small{\textcolor{darkred}{+3.8}} & 69.4 \small{\textcolor{darkred}{+0.8}} & \textbf{51.5} \small{\textcolor{darkred}{+12.4}} & \underline{40.2} \small{\textcolor{darkred}{+0.7}} & \textbf{77.4} \small{\textcolor{darkred}{+4.9}} & \textbf{40.0} \small{\textcolor{darkred}{+3.5}} \\
InternVL3.5-8B~\cite{wang2025internvl3_5} & 44.5 & 46.7 & 53.8 & 31.7 & 36.4 & 63.8 & 31.8 \\
\rowcolor{blue!5}\quad \textbf{+DynTrace} & 48.7 \small{\textcolor{darkred}{+4.2}} & 50.0 \small{\textcolor{darkred}{+3.3}} & 54.1 \small{\textcolor{darkred}{+0.3}} & \underline{48.9} \small{\textcolor{darkred}{+17.2}} & 38.4 \small{\textcolor{darkred}{+2.0}} & \underline{76.6} \small{\textcolor{darkred}{+12.8}} & 30.6 \small{\textcolor{blue}{-1.2}} \\

\bottomrule
\end{tabular}%
}
\end{table*}

\clearpage

\noindent including \textit{Camera Motion \& Orientation}, \textit{Camera-Object Interaction}, and \textit{Temporal \& Visual Changes}. The benchmark is designed to evaluate whether models can perceive, track, and reason about dynamic objects, evolving scenes, and camera motion in a unified 4D setting. 

\textit{VLM4D}~\cite{zhou2025vlm4d} focuses on spatio-temporal awareness in video-language models. The benchmark contains 1,000 videos and 1,816 QA pairs, including 600 real videos and 400 synthetic videos. Its real subset is divided into \textit{ego-centric} and \textit{exo-centric} settings, while the synthetic subset emphasizes \textit{directional} reasoning and \textit{false-positive} detection. More broadly, the benchmark is designed to probe perspective awareness, motion continuity, and motion reasoning under both real-world and synthetic conditions. 

\textit{DSI-Bench}~\cite{zhang2025dsibench} studies dynamic spatial intelligence through 943 videos and over 1,700 manually annotated VQA pairs. It decouples observer motion and object motion through three major task families, namely \textit{Object-Scene}, \textit{Observer-Scene}, and \textit{Observer-Object}, and further instantiates six question types: \textit{Fixed-Obs.}, \textit{Dyn-Obs.}, \textit{Static-Sce.}, \textit{Dyn-Sce.}, \textit{Distance}, and \textit{Orientation}. The benchmark is designed to evaluate whether models can reason about spatial changes under dynamic observer-object interactions while reducing spatial and temporal biases through symmetric sample construction.

\subsection{Further Empirical Study}

\paragraph{Fine-grained Benchmark Results.}
Tab.~\ref{tab:vlm4d_results} and Tab.~\ref{tab:dsibench_results} provide the detailed breakdowns for VLM4D and DSI-Bench. On VLM4D, gains are particularly evident in the \textit{Exo-centric} and \textit{Directional} subsets, especially for synthetic videos where visually clean trajectories allow DTV to provide highly reliable reprojected motion cues. Similarly, DSI-Bench results demonstrate substantial improvements in observer-centric categories like \textit{Static-Sce.} and \textit{Distance}, which demand the precise separation of observer motion from true object motion. Across both datasets, these consistent trends validate our core design: DTV effectively reduces motion ambiguity caused by viewpoint changes, while DTG explicitly preserves the evolving temporal traces behind each object. Ultimately, these complementary components enable DynTrace to deliver broad improvements across diverse viewpoint settings and model families, supporting robust 4D spatio-temporal reasoning.

\begin{table}[h]
  \centering
  \caption{Results on 75 manually selected videos with temporary occlusion across the three benchmarks.}
  \label{tab:block75_results}
  \small
\setlength{\tabcolsep}{6pt} 
  \begin{tabular}{lccccc}
    \toprule
    \textbf{Method} & \textbf{Avg.} & \textbf{Dyn-Bench} & \textbf{VLM4D} & \textbf{DSI-Bench} \\
    \midrule
    Baseline & 58.1\% & 62.3\% & 48.5\% & 53.3\% \\
    DynTrace & 62.0\% & 64.6\% & 56.3\% & 66.7\% \\
    \bottomrule
  \end{tabular}
\end{table}

\paragraph{Occlusion robustness.}
Tab.~\ref{tab:block75_results} evaluates 75 videos in which the main moving object is temporarily occluded and later reappears on Qwen3-VL-8B. DynTrace improves the overall accuracy from 58.1\% to 62.0\%, with particularly clear gains on VLM4D and DSI-Bench. This result supports our robustness claim: even when visible evidence is interrupted, DTG still preserves a usable dynamic trace, and DTV continues to supply geometry-informed visual priors after the target reappears.

\begin{table}[h]
\caption{Comparison on the Dyn-Bench subset with ground-truth masks and depth, where Qwen3-VL-8B uses DTV and DTG built from GT signals or from the standard DynTrace pipeline.}
\label{tab:gt_dyntrace_results}
  \small
\setlength{\tabcolsep}{4.5pt} 
\begin{tabular}{lcccc}
\toprule
\textbf{Method} & \textbf{Avg.} & \cellcolor{orange!25}\textbf{Inter-Obj.} & \cellcolor{yellow!25}\textbf{Obj.-Scene} & \cellcolor{blue!15}\textbf{Camera-Obj.} \\
\midrule
GT-Based & 55.9\% & 50.0\% & 70.8\% & 50.0\%  \\
DynTrace & 54.2\% & 47.9\% & 68.8\% & 45.8\%   \\
\bottomrule
\end{tabular}
\end{table}

\vspace{-2em}
\paragraph{Robustness to Intermediate Perception Errors.}
Tab.~\ref{tab:gt_dyntrace_results} compares two ways of building DTV and DTG for Qwen3-VL-8B on the Dyn-Bench subset with ground-truth masks and depth. In the first setting, we use the ground-truth masks and depth to construct DTV and DTG directly. In the second setting, we use the standard DynTrace pipeline to generate the same evidence from predicted masks and geometry. The GT-based evidence reaches 55.9\%, while the standard DynTrace pipeline achieves 54.2\%. This modest gap indicates that DynTrace remains robust even when its intermediate masks and geometry come from practical model predictions rather than ideal annotations.

\begin{figure}[h]
\centering
\includegraphics[width=1\columnwidth]{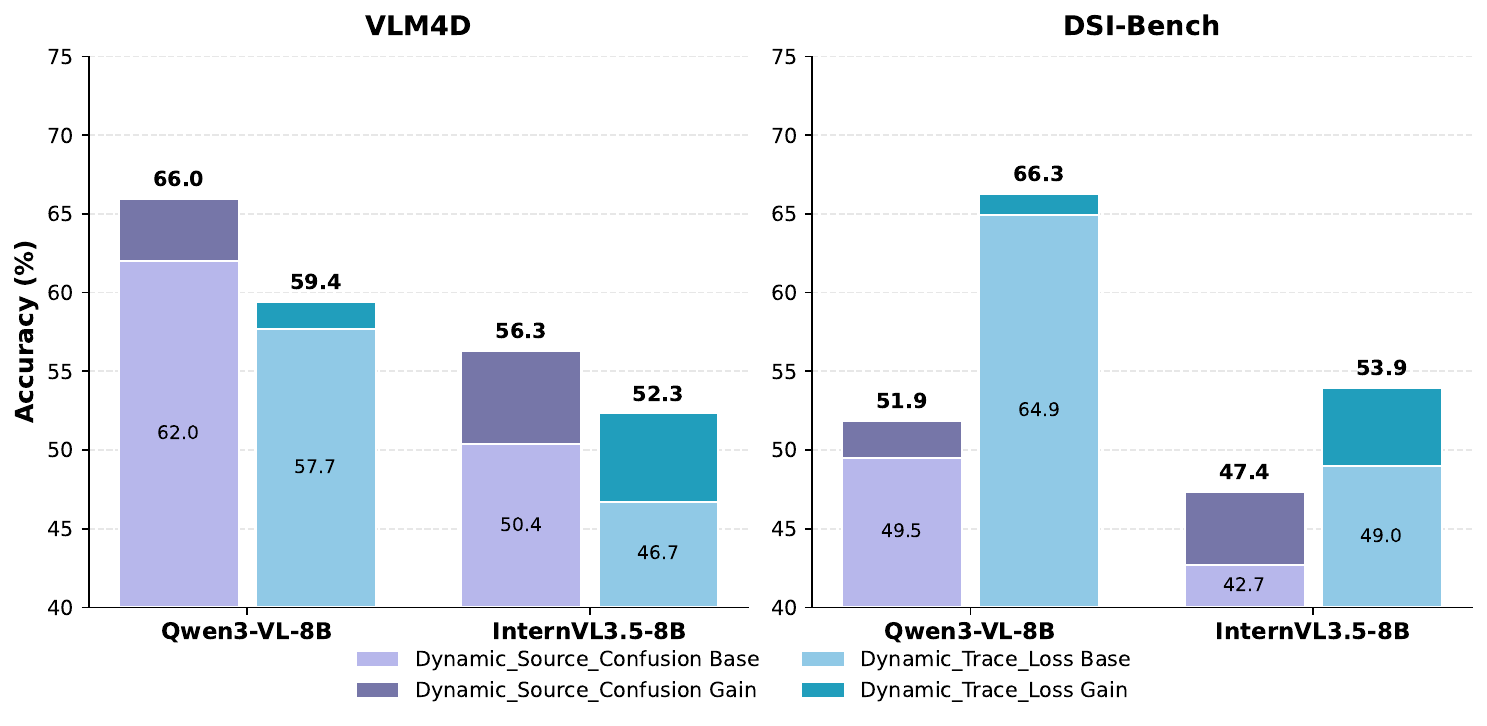}
\caption{Failure-mode analysis on VLM4D and DSI-Bench.}
\label{fig:appendix_vlm_dsi_modes}
\end{figure}

\begin{figure*}[t]
        \centering
        \includegraphics[width=0.9\textwidth]{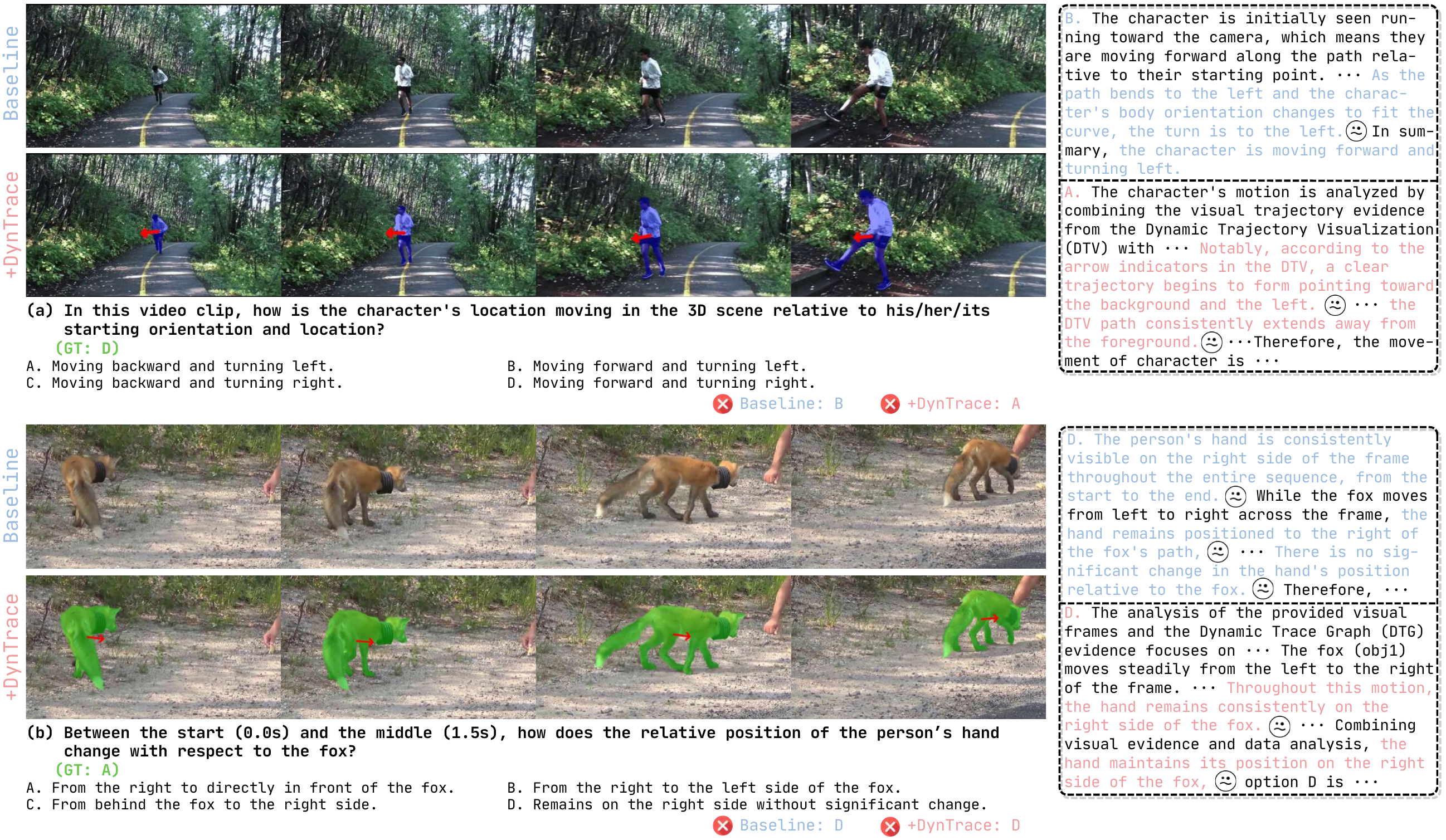}
        \caption{Representative limitation cases of DynTrace.}
        \label{fig:appendix_failures}
\end{figure*}

\vspace{-2em}
\paragraph{Failure-mode transfer.}
The exact prompt used to assign \textit{Dynamic Source Confusion} and \textit{Dynamic Trace Loss} labels is summarized in Section~\ref{app:prompt_templates}. Using the same taxonomy as in the main paper, Fig.~\ref{fig:appendix_vlm_dsi_modes} further extends the failure-mode analysis to VLM4D and DSI-Bench. On VLM4D, Qwen3-VL-8B improves from 62.0\% to 66.0\% on the \textit{Dynamic Source Confusion} subset and from 57.7\% to 59.4\% on the \textit{Dynamic Trace Loss} subset; similar gains also appear for InternVL3.5-8B. On DSI-Bench, both backbones improve on both subsets as well. These results indicate that the two failure modes are not unique to Dyn-Bench, and that DTV and DTG remain effective when transferred to other dynamic reasoning benchmarks.

\vspace{-1em}
\paragraph{Current limitations.}
Fig.~\ref{fig:appendix_failures} presents two representative cases that clarify the current boundaries of DynTrace. In case (a), the question requires converting motion into the object's own first-person frame. DynTrace already improves observer-aligned and third-person dynamic reasoning, but its current evidence remains organized around object trajectories and camera-relative context, which is less sufficient when an additional egocentric frame transformation is required for answering. In case (b), the question depends on a fine-grained local relation between a specific hand and a fox. DynTrace models the dynamic object as a whole and therefore contributes less when reasoning must rely on subtle body-part motion or part-level spatial change. These cases suggest a clear direction for future work: extending dynamic object evidence toward object-centric viewpoint conversion and finer-grained part-level motion modeling.

\section{Additional Visualizations and Prompt Templates}
\label{app:additonal}

\subsection{Additional Visualizations}

\begin{figure*}[t]
    \begin{minipage}[c][\textheight][c]{\textwidth}
        \centering
        \includegraphics[width=0.9\textwidth]{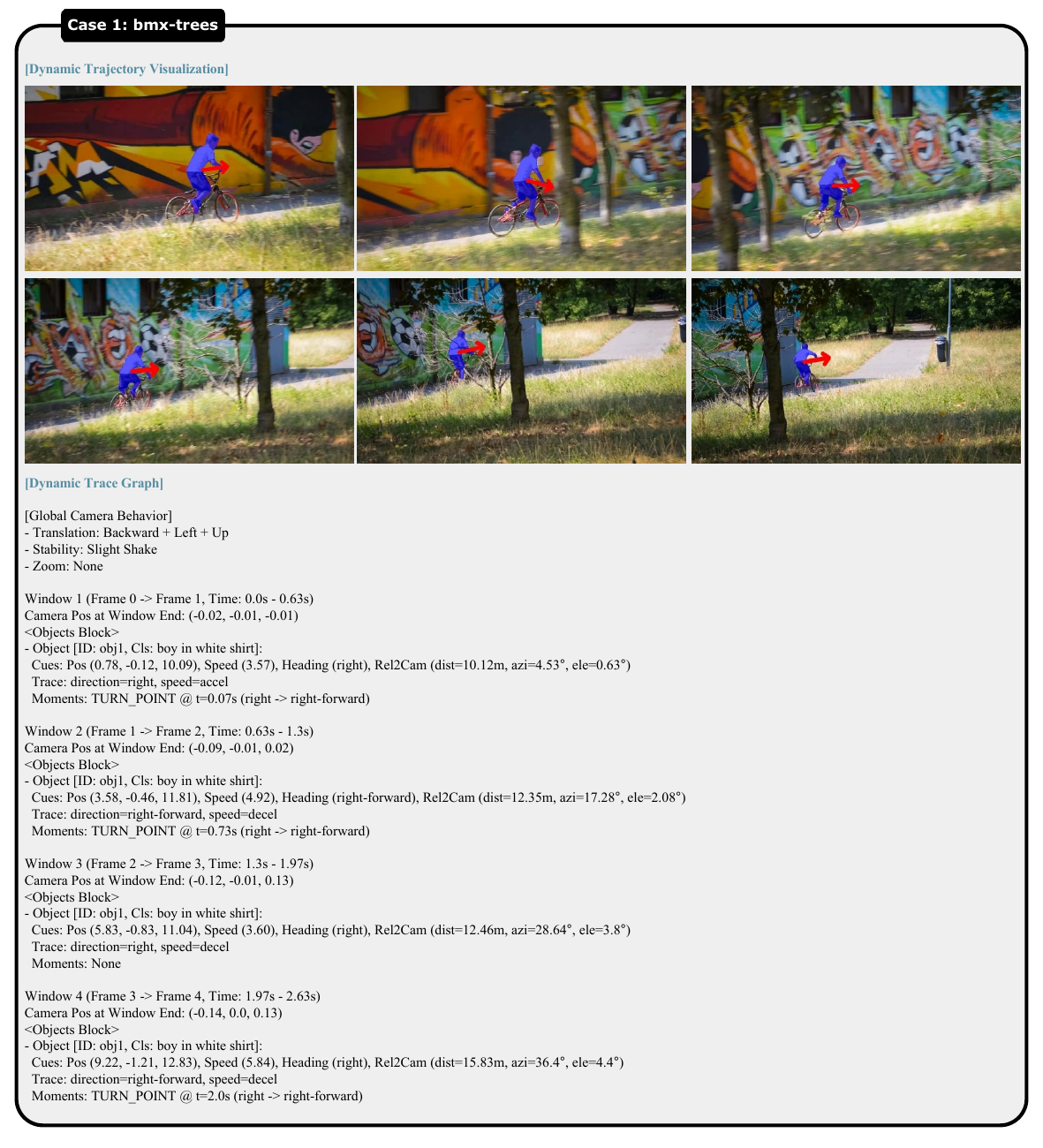}
        \caption{Complete DTV and DTG visualization for Case 1 (bmx-trees).}
        \label{fig:appendix_case1}
    \end{minipage}
\end{figure*}

\begin{figure*}[t]
    \begin{minipage}[c][\textheight][c]{\textwidth}
        \centering
        \includegraphics[width=0.9\textwidth]{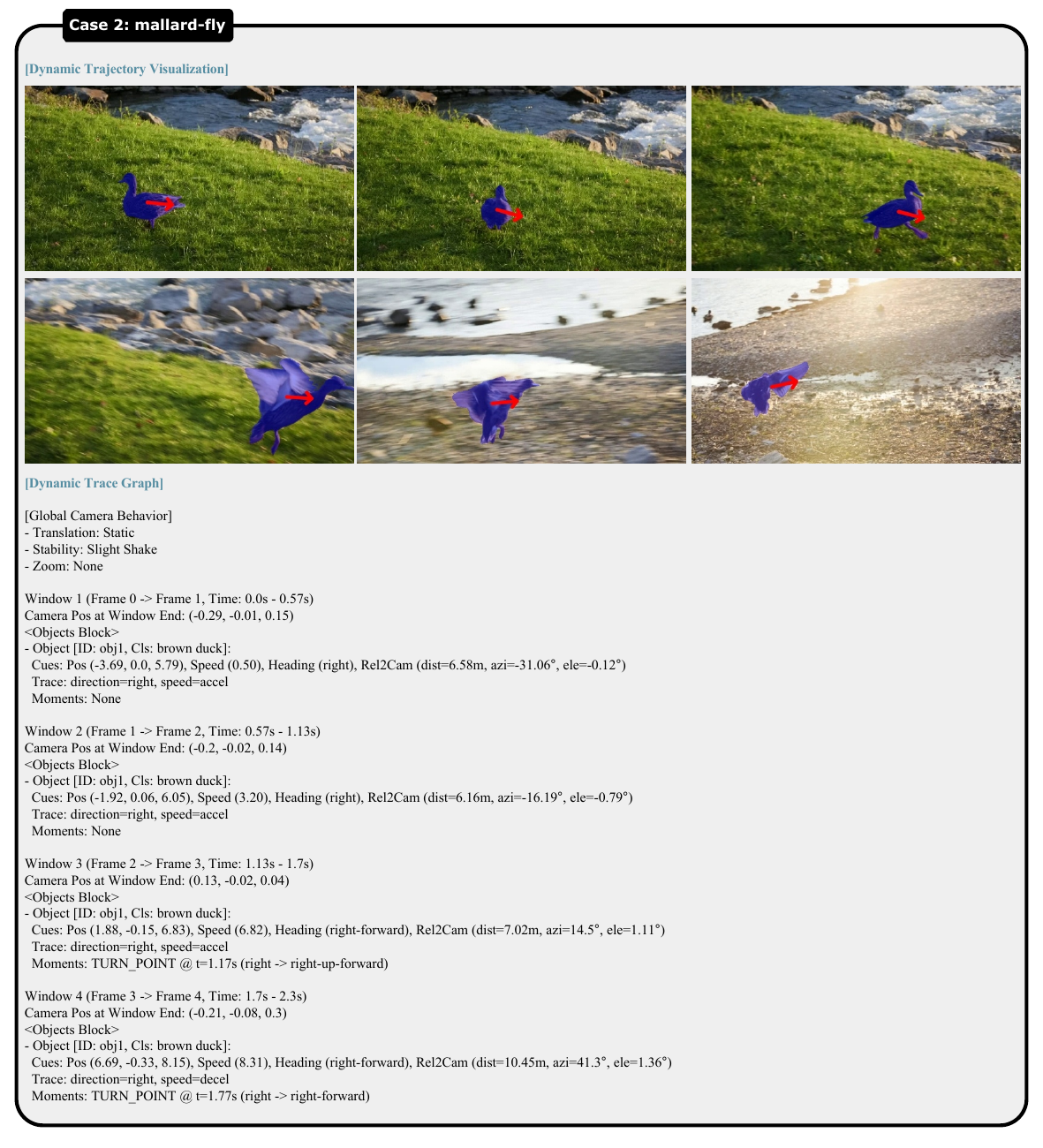}
        \caption{Complete DTV and DTG visualization for Case 2 (mallard-fly).}
        \label{fig:appendix_case2}
    \end{minipage}
\end{figure*}

\begin{figure*}[t]
    \begin{minipage}[c][\textheight][c]{\textwidth}
        \centering
        \includegraphics[width=0.9\textwidth]{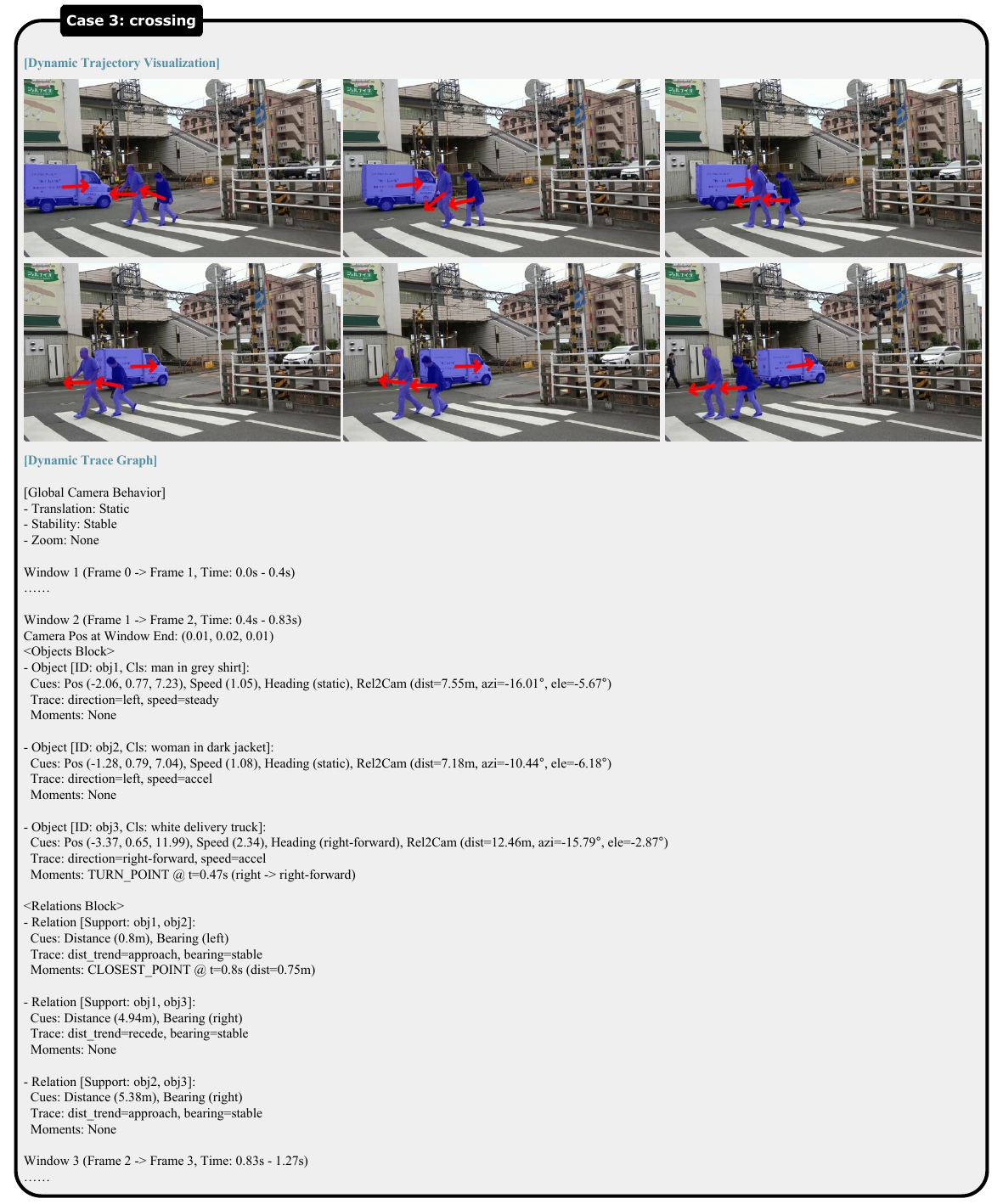}
        \caption{Complete DTV and DTG visualization for Case 3 (crossing).}
        \label{fig:appendix_case3}
    \end{minipage}
\end{figure*}

Figs.~\ref{fig:appendix_case1}, \ref{fig:appendix_case2}, and \ref{fig:appendix_case3} show the complete DTV and DTG evidence generated by DynTrace on three scenes. Case 1 highlights a single-object trajectory under a moving camera, Case 2 shows a clean long-range motion pattern with stable camera behavior, and Case 3 highlights multi-object relation evolution together with key moments such as closest approach and turning. These examples make the full evidence path visible, from geometry-informed visual priors to serialized structured traces.

\begin{figure*}[t]
    \begin{minipage}[c][\textheight][c]{\textwidth}
        \centering
        \includegraphics[width=0.9\textwidth]{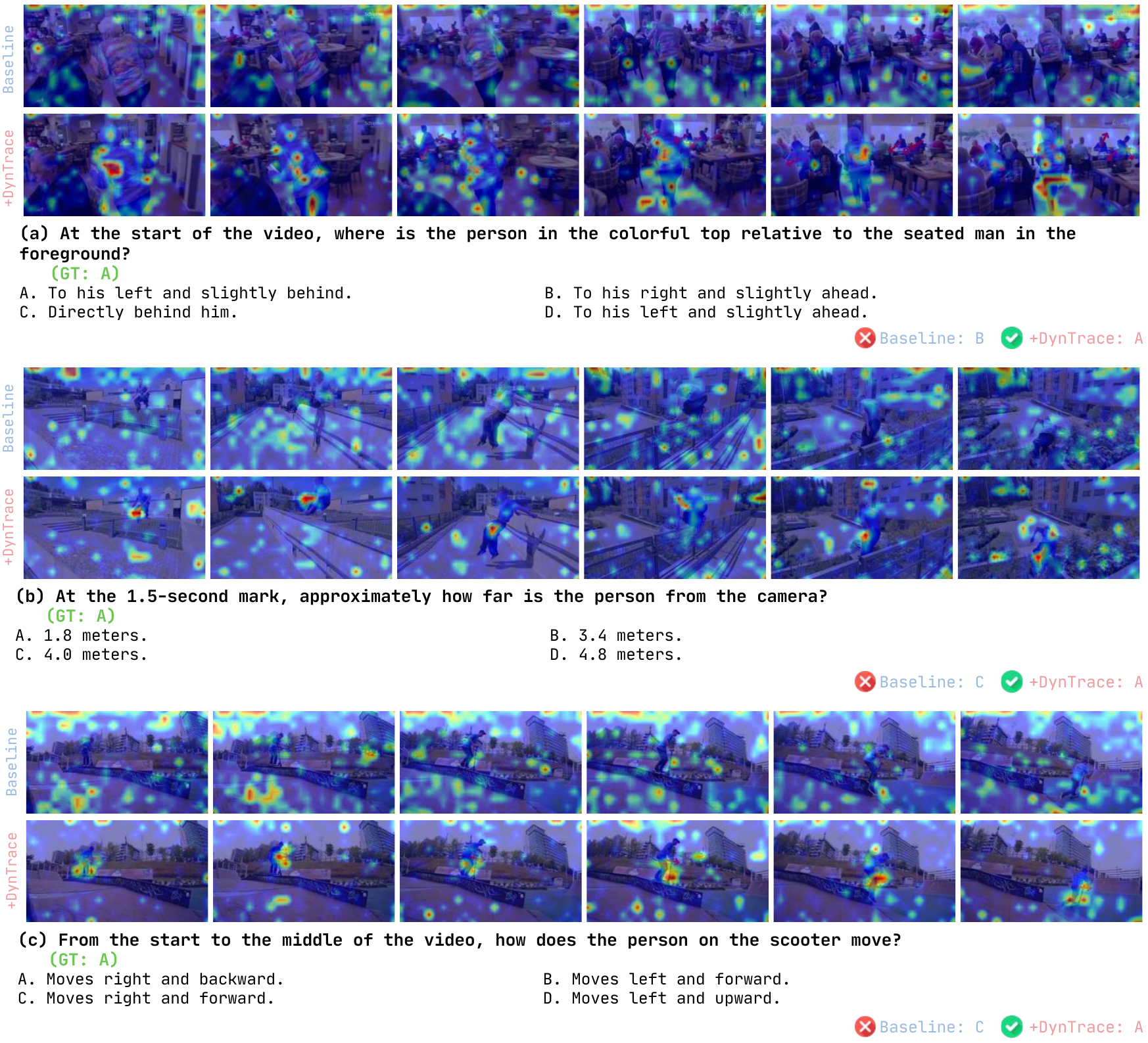}
        \caption{Attention comparison between the baseline and DynTrace on representative \textit{Dynamic Trace Loss} cases.}
        \label{fig:appendix_attention_case}
    \end{minipage}
\end{figure*}

Fig.~\ref{fig:appendix_attention_case} compares the attention behavior of the baseline and DynTrace on representative \textit{Dynamic Trace Loss} questions. The baseline attention gradually drifts toward visually salient but temporally irrelevant regions, while DynTrace keeps the attention concentrated on the query-relevant dynamic object and its evolution. This behavior is consistent with our main claim that DTG alleviates attention drift by providing an explicit temporal scaffold instead of forcing the MLLM to reconstruct the entire trace from sparse frame observations.

\begin{figure*}[p]
    \begin{minipage}[c][\textheight][c]{\textwidth}
        \centering
        \includegraphics[width=0.9\textwidth]{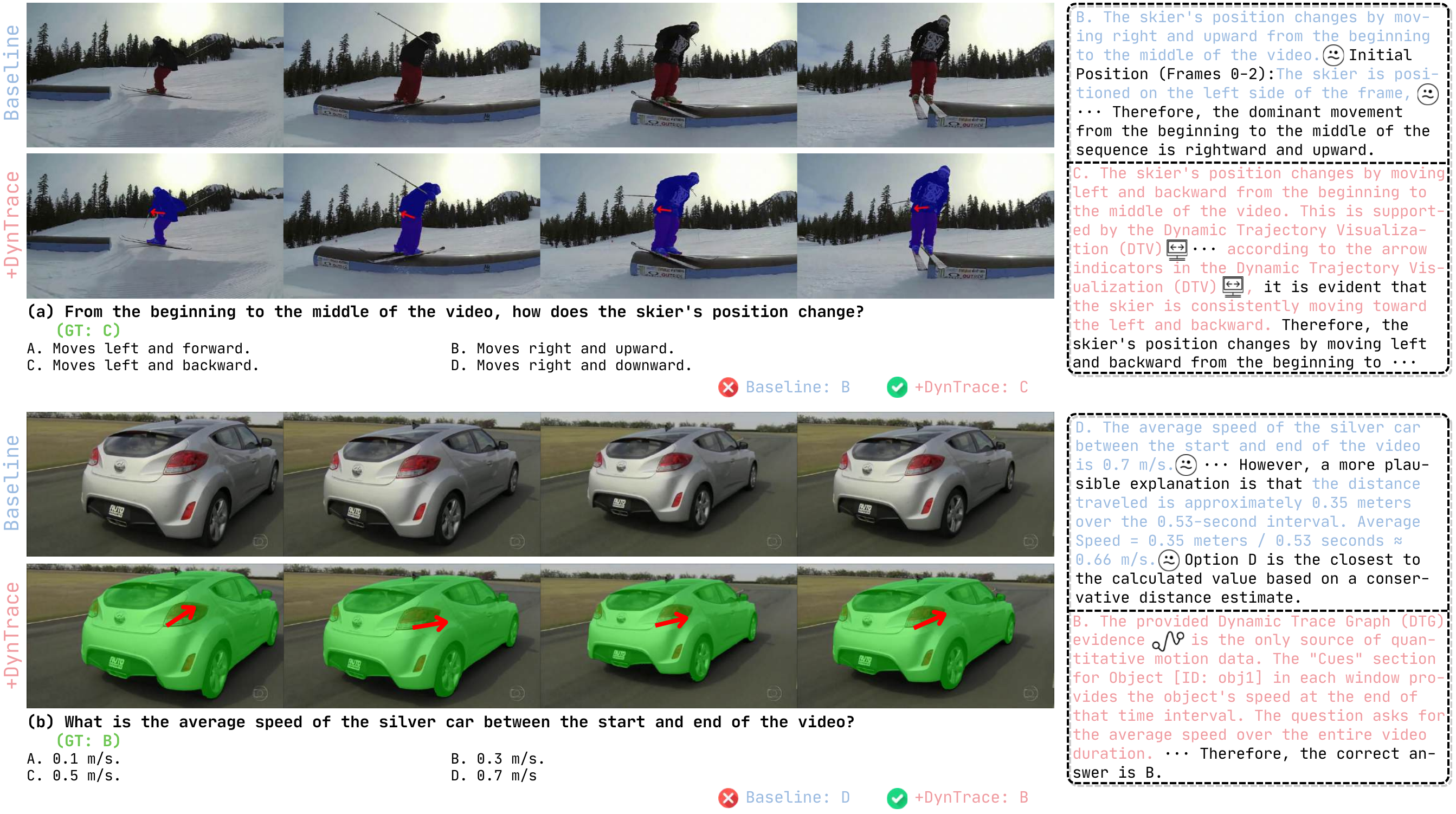}
        \caption{Additional baseline-versus-DynTrace comparisons on viewpoint-sensitive motion and metric reasoning questions.}
        \label{fig:appendix_cases1c}
        \vspace{0.5cm}
        \includegraphics[width=0.9\textwidth]{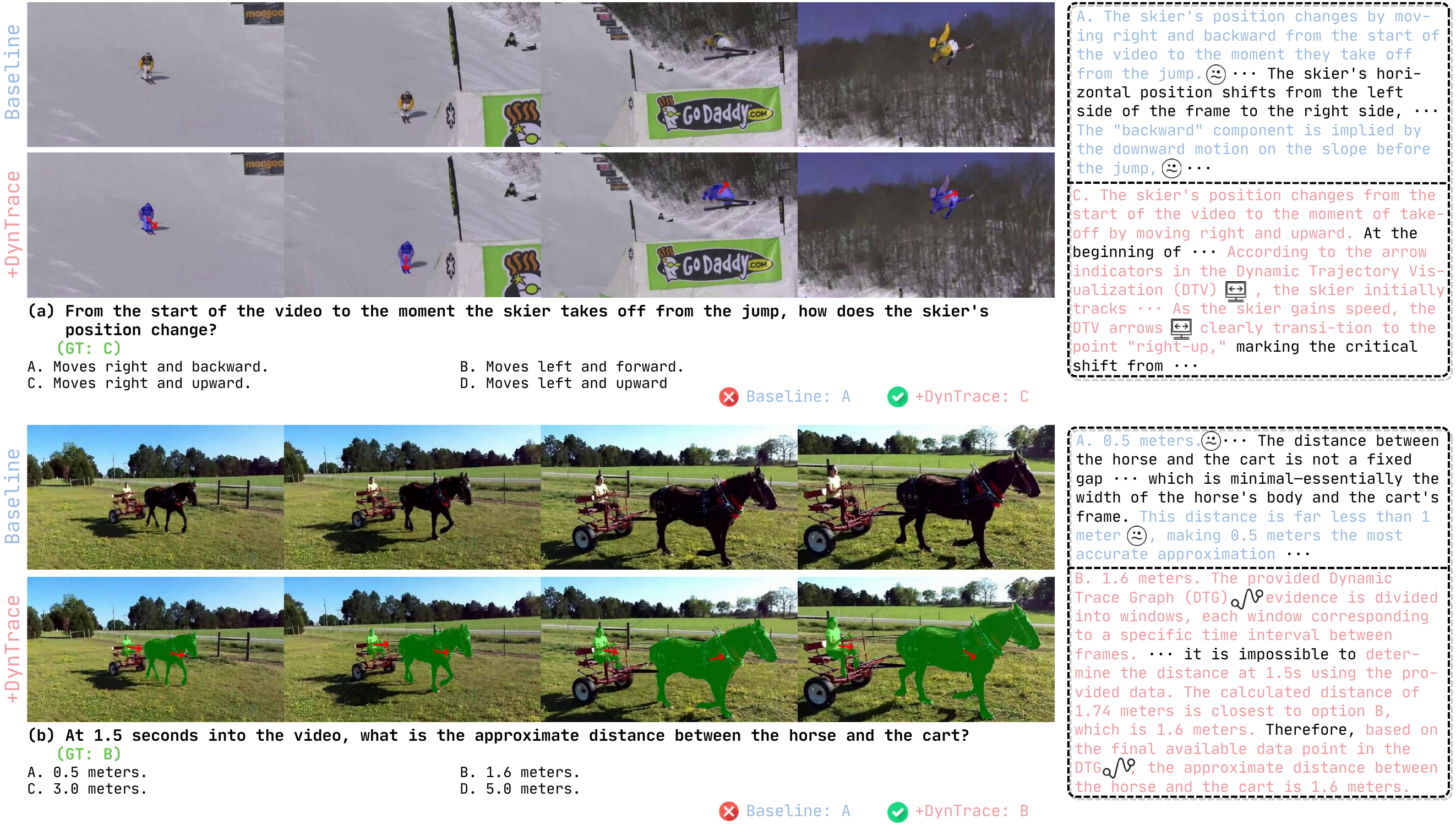}
        \caption{Additional baseline-versus-DynTrace comparisons on trajectory change and interaction distance reasoning questions.}
        \label{fig:appendix_cases2c}
    \end{minipage}
\end{figure*}

Figs.~\ref{fig:appendix_cases1c} and \ref{fig:appendix_cases2c} provide additional question-level comparisons between the baseline and DynTrace. They cover viewpoint-sensitive motion direction, metric speed estimation, trajectory change before takeoff, and interaction-distance reasoning. Across these examples, DTV provides the visual prior needed to correct ambiguous motion interpretation, while DTG exposes explicit trace evolution and key moments that support more reliable answer selection.

\subsection{Prompt Templates}
\label{app:prompt_templates}
\begin{figure*}[t]
    \begin{minipage}[c][\textheight][c]{\textwidth}
        \centering
        \includegraphics[width=0.75\textwidth]{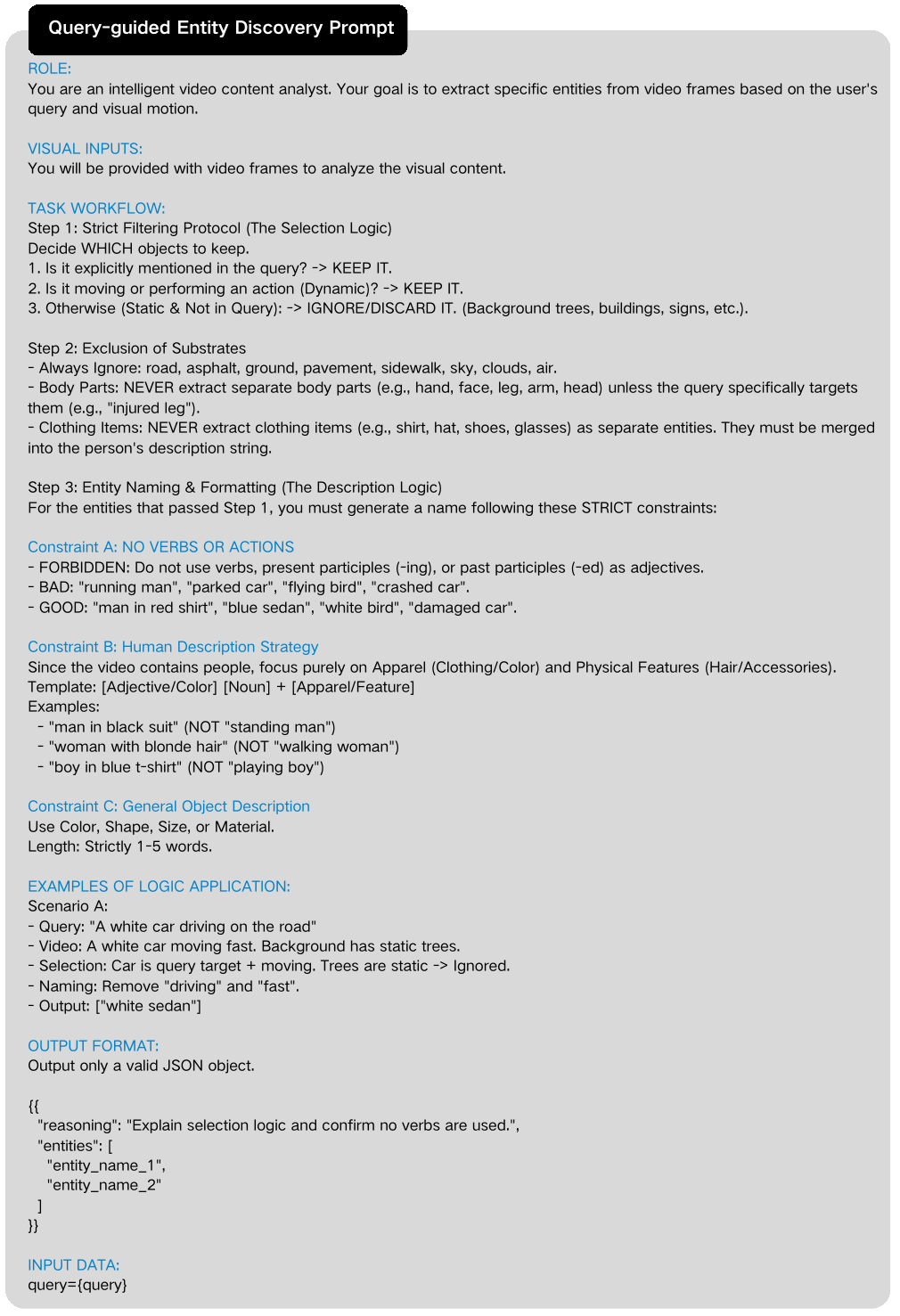}
        \caption{Prompt template for query-guided entity discovery.}
        \label{fig:entity}
    \end{minipage}
\end{figure*}

\begin{figure*}[t]
    \begin{minipage}[c][\textheight][c]{\textwidth}
        \centering
        \includegraphics[width=0.75\textwidth]{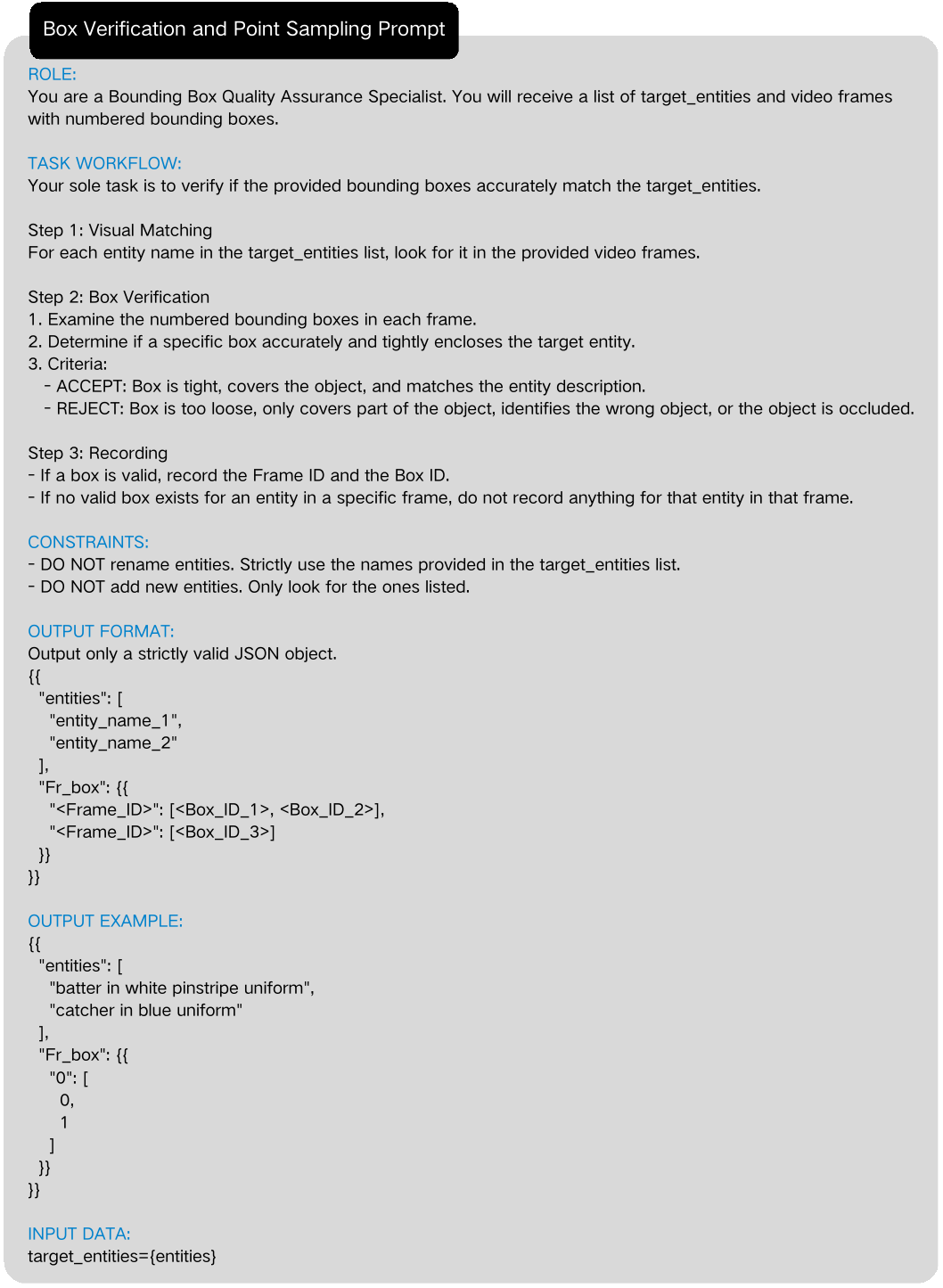}
        \caption{Prompt template for multi-frame box verification and retained-box selection.}
        \label{fig:box}
    \end{minipage}
\end{figure*}

\begin{figure*}[t]
    \begin{minipage}[c][\textheight][c]{\textwidth}
        \centering
        \includegraphics[width=0.9\textwidth]{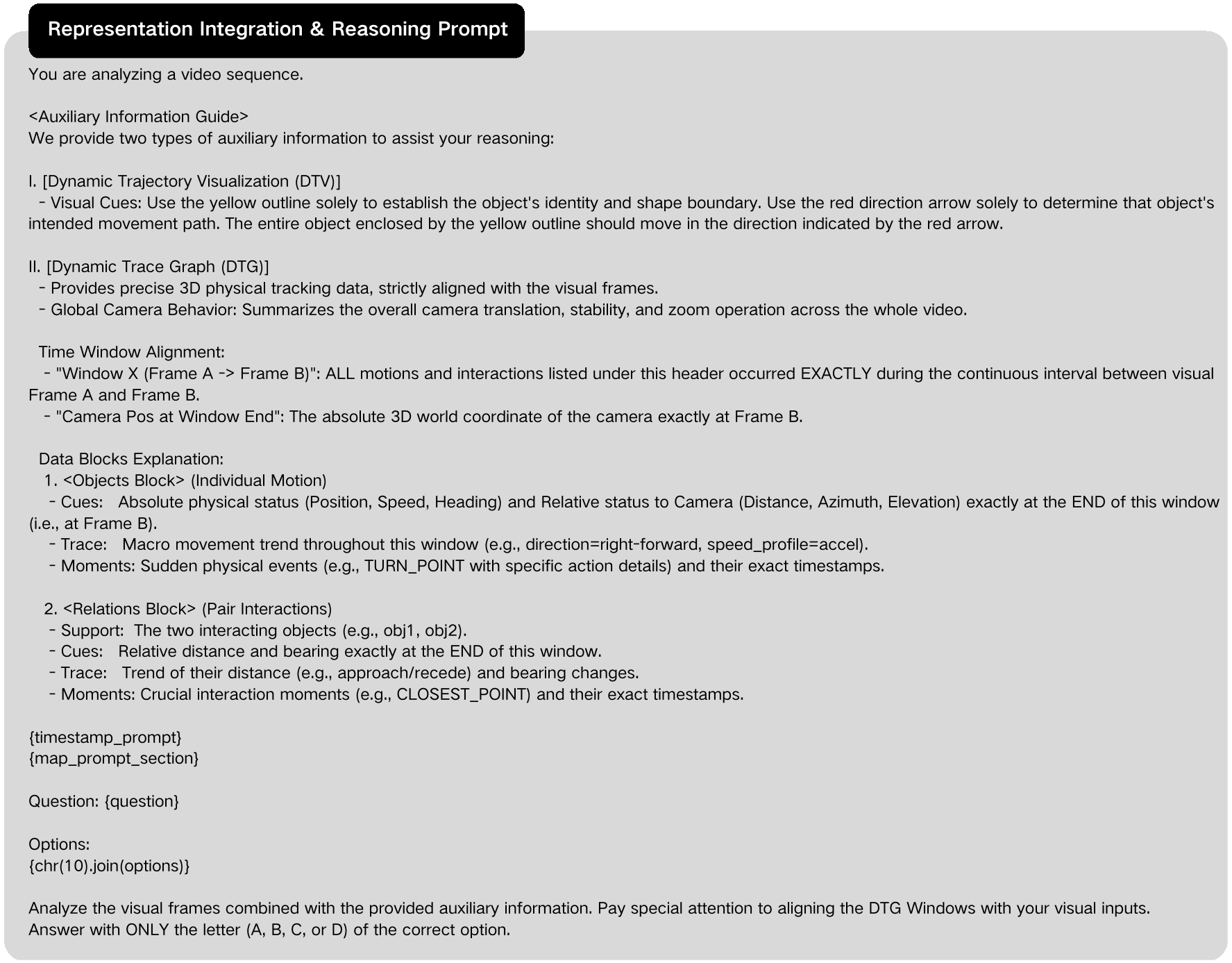}
        \caption{Prompt template for Representation Integration \& Reasoning.}
        \label{fig:appendix_stage3_prompt}
    \end{minipage}
\end{figure*}

\begin{figure*}[t]
    \begin{minipage}[c][\textheight][c]{\textwidth}
\centering
\includegraphics[width=0.9\textwidth]{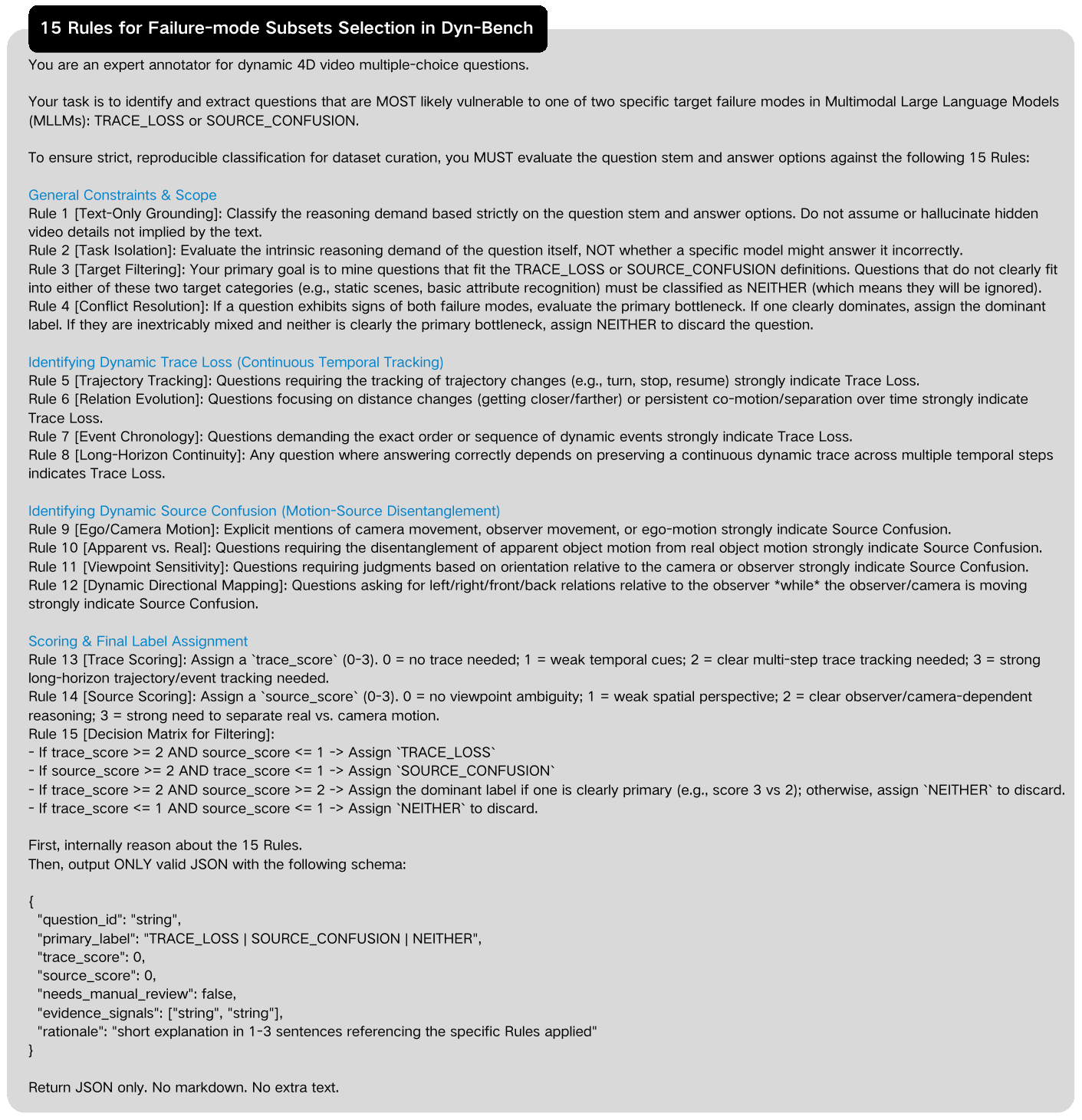}
\caption{Prompt template for the 15-rule failure-mode selection.}
\label{fig:appendix_15rules}
    \end{minipage}
\end{figure*}

This section collects four prompt templates used by \textbf{DynTrace}. Fig.~\ref{fig:entity} presents the query-guided entity discovery prompt for extracting the initial set of dynamic entities from the input video and question. Fig.~\ref{fig:box} shows the multi-frame box verification prompt used to validate candidate motion boxes before point sampling. Fig.~\ref{fig:appendix_stage3_prompt} shows the final reasoning prompt that combines the user query with DTV and DTG. Fig.~\ref{fig:appendix_15rules} provides the rule-based prompt used to classify questions into the \textit{Dynamic Source Confusion} and \textit{Dynamic Trace Loss} subsets for the failure-mode analysis.

\clearpage

\clearpage

\clearpage

\clearpage

\clearpage

\clearpage

\clearpage

\clearpage

\clearpage

\clearpage

\end{document}